\documentclass[aip,amsmath,amssymb,reprint]{revtex4-1}
\usepackage{graphicx}
\usepackage{dcolumn}
\usepackage{bm}
\usepackage[utf8]{inputenc}
\usepackage[T1]{fontenc}
\usepackage{mathptmx}
\usepackage{etoolbox}
\usepackage{orcidlink}
\usepackage{CJKutf8}

\usepackage{subfloat}
\usepackage{caption} 
\usepackage{subcaption}

\makeatletter
\def\@email#1#2{%
 \endgroup
 \patchcmd{\titleblock@produce}
  {\frontmatter@RRAPformat}
  {\frontmatter@RRAPformat{\produce@RRAP{*#1\href{mailto:#2}{#2}}}\frontmatter@RRAPformat}
  {}{}
}%
\makeatother
\begin{document}
\begin{CJK*}{UTF8}{gbsn}
\preprint{AIP/123-QED}

\title{Quantifying patterns of punctuation in modern Chinese prose}

\author{Michał Dolina \orcidlink{0000-0001-8980-9727}}
\affiliation{Faculty of Computer Science and Telecommunications, Cracow University of Technology, ul.~Warszawska 24, 31-155 Krak\'ow, Poland}

\author{Jakub Dec \orcidlink{0000-0001-9946-1541}} 
\affiliation{Faculty of Computer Science and Telecommunications, Cracow University of Technology, ul.~Warszawska 24, 31-155 Krak\'ow, Poland}

\author{Stanisław Drożdż \orcidlink{0000-0003-1613-6175}}
 \email{stanislaw.drozdz@ifj.edu.pl}
\affiliation{Faculty of Computer Science and Telecommunications, Cracow University of Technology, ul.~Warszawska 24, 31-155 Krak\'ow, Poland}
\affiliation{Complex Systems Theory Department, Institute of Nuclear Physics, Polish Academy of Sciences, ul.~Radzikowskiego 152, 31-342 Krak\'ow, Poland }

\author{Jarosław Kwapień \orcidlink{0000-0001-8813-9637}}
\affiliation{Complex Systems Theory Department, Institute of Nuclear Physics, Polish Academy of Sciences, ul.~Radzikowskiego 152, 31-342 Krak\'ow, Poland }

\author{Jin Liu \orcidlink{0000-0001-6757-8008 }}
\affiliation{School of Modern Languages, Georgia Institute of Technology, Atlanta, GA, USA}
 
\author{Tomasz Stanisz \orcidlink{0000-0002-4787-6881}}
\affiliation{Complex Systems Theory Department, Institute of Nuclear Physics, Polish Academy of Sciences, ul.~Radzikowskiego 152, 31-342 Krak\'ow, Poland }

\date{\today}

\begin{abstract}
Recent research shows that punctuation patterns in texts exhibit universal features across languages. Analysis of Western classical literature reveals that the distribution of spaces between punctuation marks aligns with a discrete Weibull distribution, typically used in survival analysis. By extending this analysis to Chinese literature represented here by three notable contemporary works, it is shown that Zipf’s law applies to Chinese texts similarly to Western texts, where punctuation patterns also improve adherence to the law. Additionally, the distance distribution between punctuation marks in Chinese texts follows the Weibull model, though larger spacing is less frequent than in English translations. Sentence-ending punctuation, representing sentence length, diverges more from this pattern, reflecting greater flexibility in sentence length. This variability supports the formation of complex, multifractal sentence structures, particularly evident in Gao Xingjian’s \textit{Soul Mountain}. These findings demonstrate that both Chinese and Western texts share universal punctuation and word distribution patterns, underscoring their broad applicability across languages.
\end{abstract}

\maketitle

\begin{quotation}

In the realm of complex systems~\cite{KwapienJ-2012a}, featuring characteristics such as multilevel hierarchical organization, long-range correlations, and an absence of characteristic scale, the Chinese language introduces its own unique complexities. Language, regardless of its specific nature, serves as a system where complexity prominently manifests itself. It intricately weaves together seemingly simple elements into structures capable of articulating an almost infinite array of concepts at varying levels of sophistication. The quantitative analysis of language, encompassing investigations into the statistical properties of texts, aims to unveil correlations and even laws delineating the measurable properties of language and the underlying processes governing their manifestation. Such an approach becomes more complete when examining the intricacies of the Chinese language, whose written form has the world's longest history of continuous use and remains in practice today, and its complexity lies in various aspects, including its writing system, tonal nature, and cultural nuances. In particular, the role of punctuation in Chinese linguistics adds an additional layer of intricacy to the study of language systems. Understanding the related linguistic processes, with a specific focus on the punctuation in Chinese, holds immense potential for shedding light on still unanswered questions in linguistic research. This includes inquiries into the origins of language, the mechanisms behind its acquisition, and its representation within the human brain. Furthermore, the insights gained from such research have the power to enhance tools utilized in the field of natural language processing, a domain currently heavily influenced by deep learning and, notably, large language models (LLMs)~\cite{ShanahanM-2023a}.

\end{quotation}

\section{Introduction}

Punctuation is a fundamental aspect of written language that enhances clarity, meaning, and expression. It consists of various symbols and marks that are used to organize and structure sentences, indicating pauses, emphasis, and relationships between words and ideas~\cite{ChafeW-1988a}. The proper use of punctuation is crucial for effective communication, as it helps convey the intended message accurately and ensures that readers interpret the text as intended. As recent quantitative studies~\cite{StaniszT-2024a,DecJ-2024a} show, punctuation marks, alongside words, are subject to Zipf's law~\cite{ZipfGK-1935a}, and the distributions of distances between consecutive punctuation marks generally adhere to the discrete Weibull distribution appearing in survival analyses~\cite{WeibullW-1951a,NakagawaT-1975a,MillerJrRG-1998a}. This rule is so strong that it applies even to experimental texts in which the author uses punctuation in an unconventional way~\cite{DecJ-2024a,StaniszT-2024b}. Furthermore, it is primarily the distributions of punctuation that prove to be responsible for building long-range correlations in texts.

In this context, the intriguing question is how such characteristics are presented in the Chinese punctuation system, where punctuation has a unique origin and developmental history and has evolved over the centuries. The basic unit of writing in Chinese is the character. Unlike Western languages where words are typically made up of letters, Chinese words are composed of one or more characters. As Chinese language is morphosyllabic, characters can represent both morphemes (meaningful units) and syllables. Some characters stand alone as complete words, while others combine to form compound words. As a result, the need for a system of punctuation to guide reading and indicate grammatical structure became apparent. Rudimentary punctuation marks appeared in oracle bones and bronze inscriptions, such as the symbol ㇄ marking the end of a sentence in an inscription dated around 900 BCE~\cite{GuanX-2002a}. By the Han Dynasty (202 BCE -- 220 AD), there were approximately 15 punctuation symbols, including one that is identical to the modern-day comma (",").  The term "ju dou" ("sentence-pause") also emerged during the Eastern Han period, emphasizing the importance of segmentation in the study of Confucian classics. By the late imperial period, approximately fifty to sixty punctuation marks had accumulated, including the pause mark "、”, the circle mark "。” and the title mark《》, yet there were no equivalents to Western symbols such as the exclamation mark "!" or the question mark "?". Nevertheless, the use of traditional Chinese punctuation marks was not standardized, resulting in widespread variation. For example, some texts used the symbol "、" or "，" to function as a modern-day comma, and use "。" to indicate the end of a sentence, while others relied solely on "、" or "。" to indicate pauses, without distinguishing between commas and periods. Moreover, most official texts in Classical Chinese lacked punctuation marks, leaving readers to determine the segmentation on their own. Many traditional Chinese punctuation marks, rather than occupying a character space within the text lines, were placed in the margins and interlinear spaces of the page, functioning more as external annotations and commentary. The introduction of Western-style punctuation in the late nineteenth century and the early twentieth century had a profound impact on the punctuation system in modern China. The promotion of Western-style punctuation was articulated by Hu Shih, who was educated at Cornell and Columbia and became a prominent leader of the vernacular Chinese movement, in a 1916 article published in Kexue (Science, 1915-1950), the first periodical in modern China to adopt Western punctuation marks~\cite{ForsterE-2018a}. Throughout the 1920s and 1930s, Chinese writers gradually adopted the newly imported punctuation marks, which became integrated into the body of the text~\cite{MullaneyTS-2017a}.

Nevertheless, old habits die hard. The historical lack of clear distinctions between commas and periods in Chinese texts may explain why Chinese writers today are prone to write run-on sentences or clauses, marked only by commas or semicolons. As the segmentation of sentences in Chinese is still an unsettled and controversial issue in natural language processing, and there is no standard or a widely accepted criteria yet to define a sentence based on punctuation marks~\cite{LiuJ-2023a}, this study considers the following symbols as sentence termination characters: 。(ideographic full stop), \textbf{！}(fullwidth exclamation mark), \textbf{？}(fullwidth question mark), \textbf{…} (horizontal ellipsis), \textbf{；}(fullwidth semicolon), and \textit{newline} if symbol sequence wasn't terminated by any of the already-listed marks. Additionally, Chinese writers often use parentheses (【 】) and various hyphens to mark pauses, quotations, or emphasis. These were not included in the punctuation set studied here. There were also characters such as single angle brackets (〈 〉) or typical "Chinese" quotation marks in the form of corner brackets (「 」), which appear in the written language, but not in the source files we analyzed, in which the writers decided to use Western types of quotation marking. Proper placement and usage of these punctuation marks are essential for clarifying the intended meaning of sentences, ensuring effective communication in written Chinese. While the meaning of text is not particularly crucial in the present research, it can serve as guidance in proper dataset segmentation, however. In this way, a similar analysis of the distribution statistics of distances between consecutive punctuation marks in Chinese language texts can be systematically conducted, similar to that for Western languages. 

 A text-based dataset in "western" (Latin, alphabetical) languages can be easily interpreted as hierarchical structure, where the entire text is divided into paragraphs by newlines, paragraphs are divided into words by spaces, and finally words are divided into letters. With the contemporary Chinese punctuation, we consider “a sentence” to be a set of symbols between two sentence termination symbols. When using commas, we can also distinguish sub-sentences inside complex sentences, just like in languages that use the Latin alphabet. In Western languages, every two words are separated by a space $--$ to count words in a sentence, one can simply count the spaces (and add 1, to be precise). In Chinese, there is no such thing as `space' $-–$ there are no delimiters between words. Practical solutions for dealing with this kind of issue are provided by existing programming libraries that contain ready-to-use functions. One of them, used in the present study, is called Jieba~\cite{Jieba}.

\section{Methodology and data}

\subsection{Books studied}

In order to make the issue of identification of punctuation transparent in the present exploratory work, the following three novels representing the contemporary Chinese literature are systematically analysed: \\
(i) \textit{The Drunkard}, authored by Liu Yi-Chang and first published in 1962 as a serial in a Hong Kong evening paper. It is considered one of the first full-length stream-of-consciousness novels written in Chinese~\cite{Drunkard-CN}. The first English translation of this novel done by C.~Yiu is also included in the present analysis~\cite{Drunkard-EN}.\\
(ii) \textit{The Sun Shines over the Sanggan River}, authored by Ding Ling in 1948, drawing from her own experiences to depict the lives of peasants and the class struggles during the implementation of land reform in northern China~\cite{SangganRiver-CN}. The book was translated into English by W.J.F.~Jenner and this translation is also analyzed here~\cite{SangganRiver-EN}.\\
(iii) \textit{The Soul Mountain} written by Gao Xingjian, the recipient of the 2000 Nobel Prize in Literature. The novel, originally published in 1990, is renowned for its unique narrative style and profound exploration of identity, nature, and the human condition~\cite{SoulMountain-CN}. Its English translation of this novel by M.~Lee is also available and included~\cite{SoulMountain-EN}. 

\subsection{Zipf's law}

The most basic and fairly universal quantitative linguistic law in Western languages is Zipf's law~\cite{PiantadosiST-2014a}. The essence of this law is reflected by a power-law distribution, which states that the frequency of any given word is inversely proportional to its rank in the frequency table~\cite{ZipfGK-1935a}. In formal terms, the probability $P$ of encountering the $R$-th most frequent word scales approximately according to 
\begin{equation}
P(R) \sim R^{-\gamma}, 
\end{equation}
where $\alpha$ is typically close to 1~\cite{ZipfGK-1935a,KwapienJ-2010a}. In simpler terms, the most common word in a text occurs approximately twice as often as the second most common word, three times as often as the third most common word, and so forth. Expressed in the log-log scale this dependence converts into the straight line whose slope equals $-\gamma$. According to a model proposed by Zipf, such a dependence of frequency upon rank comes from a word-usage optimization~\cite{ZipfGK-1949a}. Recent results based on texts written in Western languages show that the power-law relation is preserved and even improved in terms of its accuracy if punctuation is also included on par with words~\cite{KuligA-2017a,StaniszT-2024a}. 

In Chinese texts, a concept of the Zipf law must be reformulated in order to deal with the absence of inter-word spaces. It was shown that the individual characters did not exhibit a power-law behavior but, if the characters were replaced by $n$-grams consisting of $n$ consecutive characters that form meaningful words, the Zipf relation could be restored~\cite{HaLQ-2003a,XiaoH-2008a}. However, the related studies did not consider punctuation marks. In light of the above remarks regarding the specificity of the concept of a word in the Chinese language, particularly with respect to punctuation and its role, it is entirely justified to begin the current analysis by verifying to what extent the law applies to Chinese texts and how punctuation functions in this context.

\begin{figure*}
\centering
\includegraphics[width=0.48\textwidth]{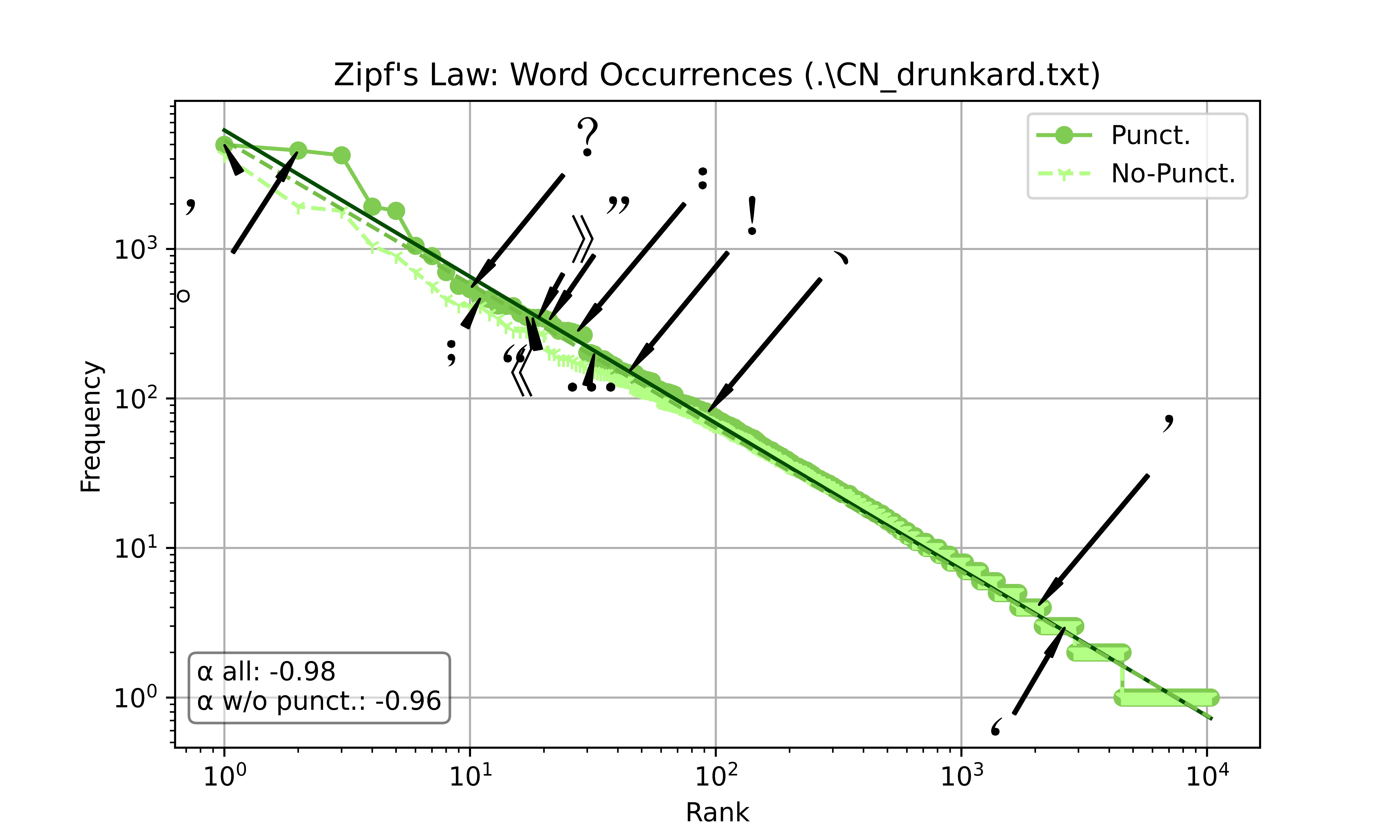}
\includegraphics[width=0.48\textwidth]{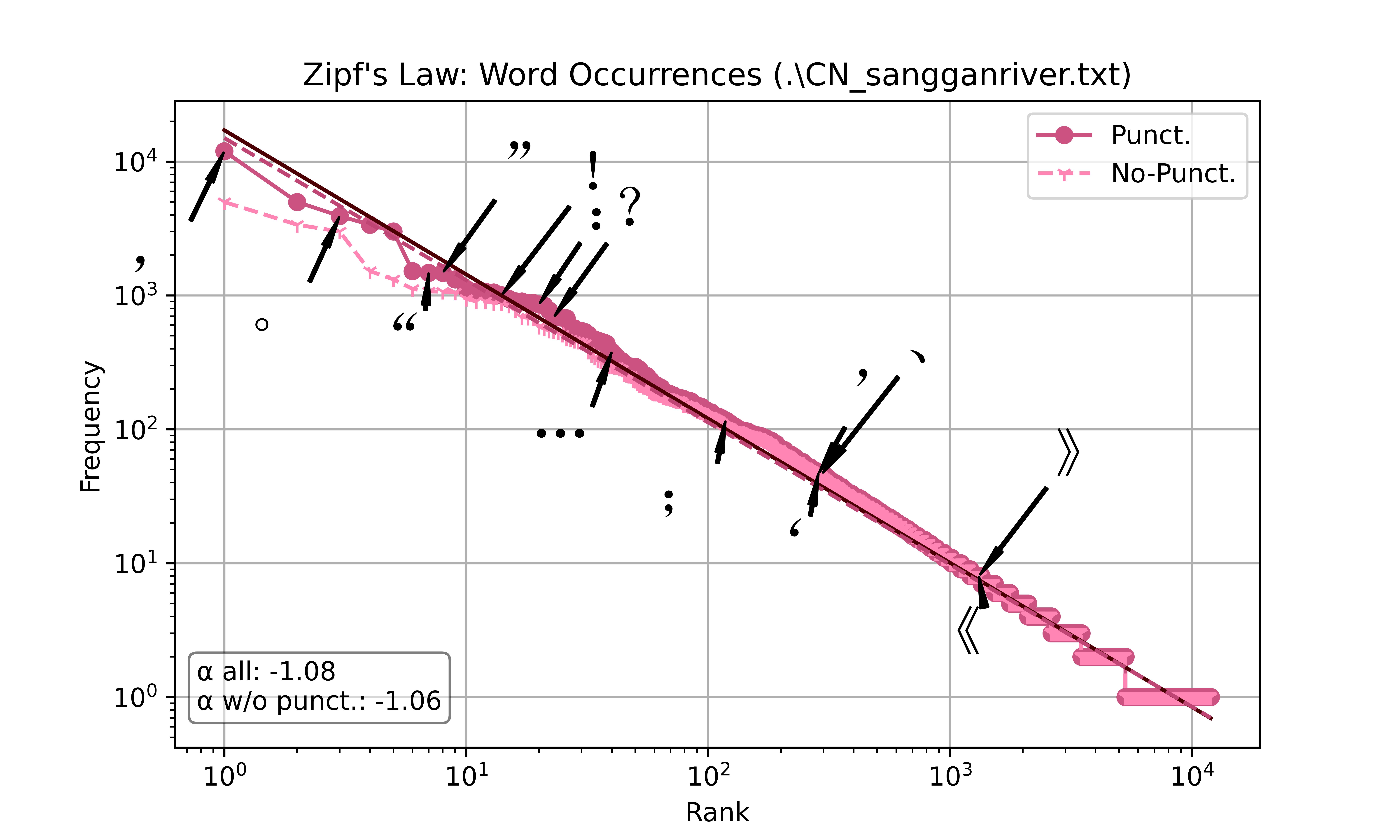}
\includegraphics[width=0.48\textwidth]{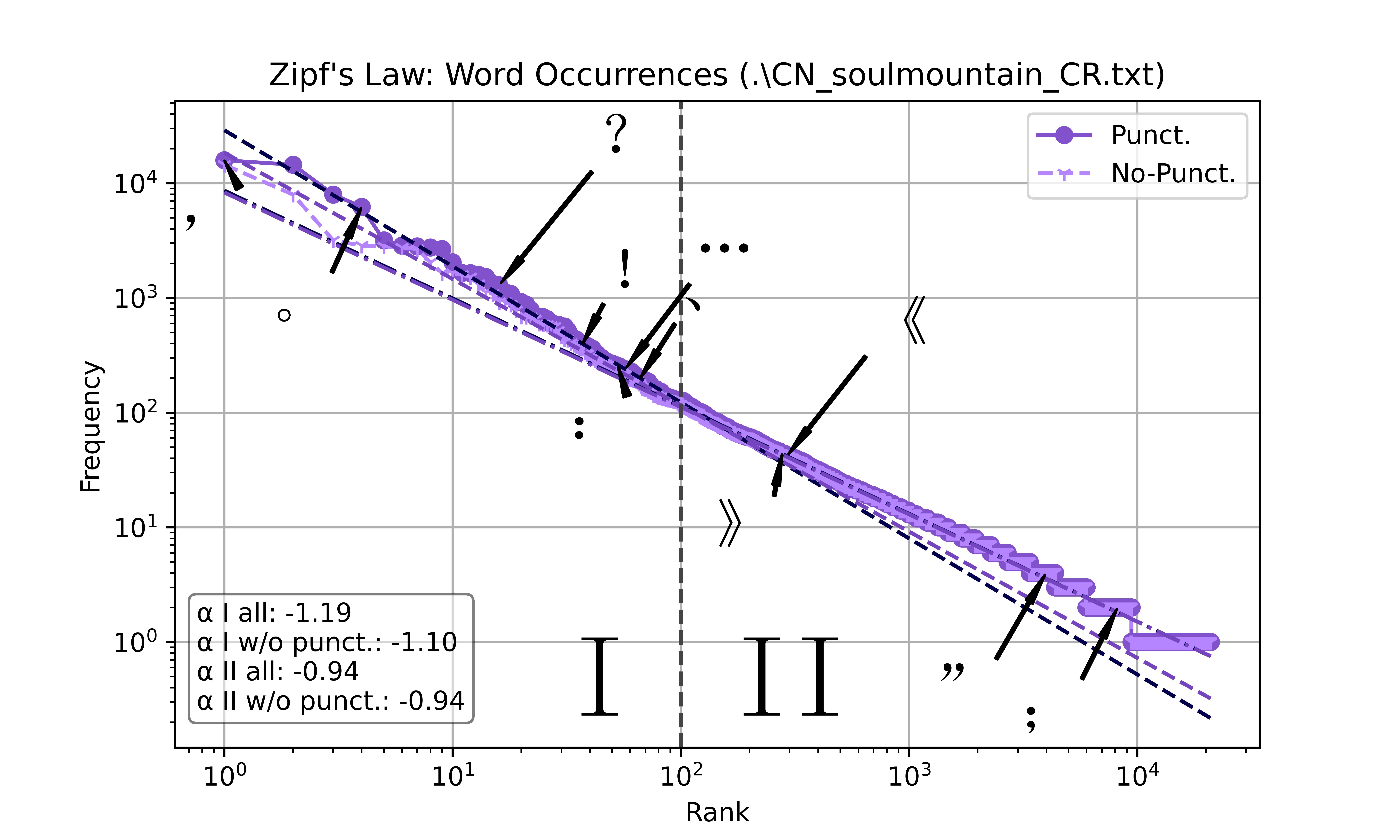}
\caption{Rank-frequency distributions of words (crosses) and words + punctuation marks (full circles) for the three Chinese books: \textit{The Drunkard} (top left), \textit{The Sun Shines over the Sanggan River} (top right), and \textit{The Soul Mountain} (bottom). The distributions are fitted with a power-law function whose scaling index $\gamma$ is given explicitly in each panel.}
\label{fig::Zipf_CN}
\end{figure*}

\begin{figure*}
\centering
\includegraphics[width=0.48\textwidth]{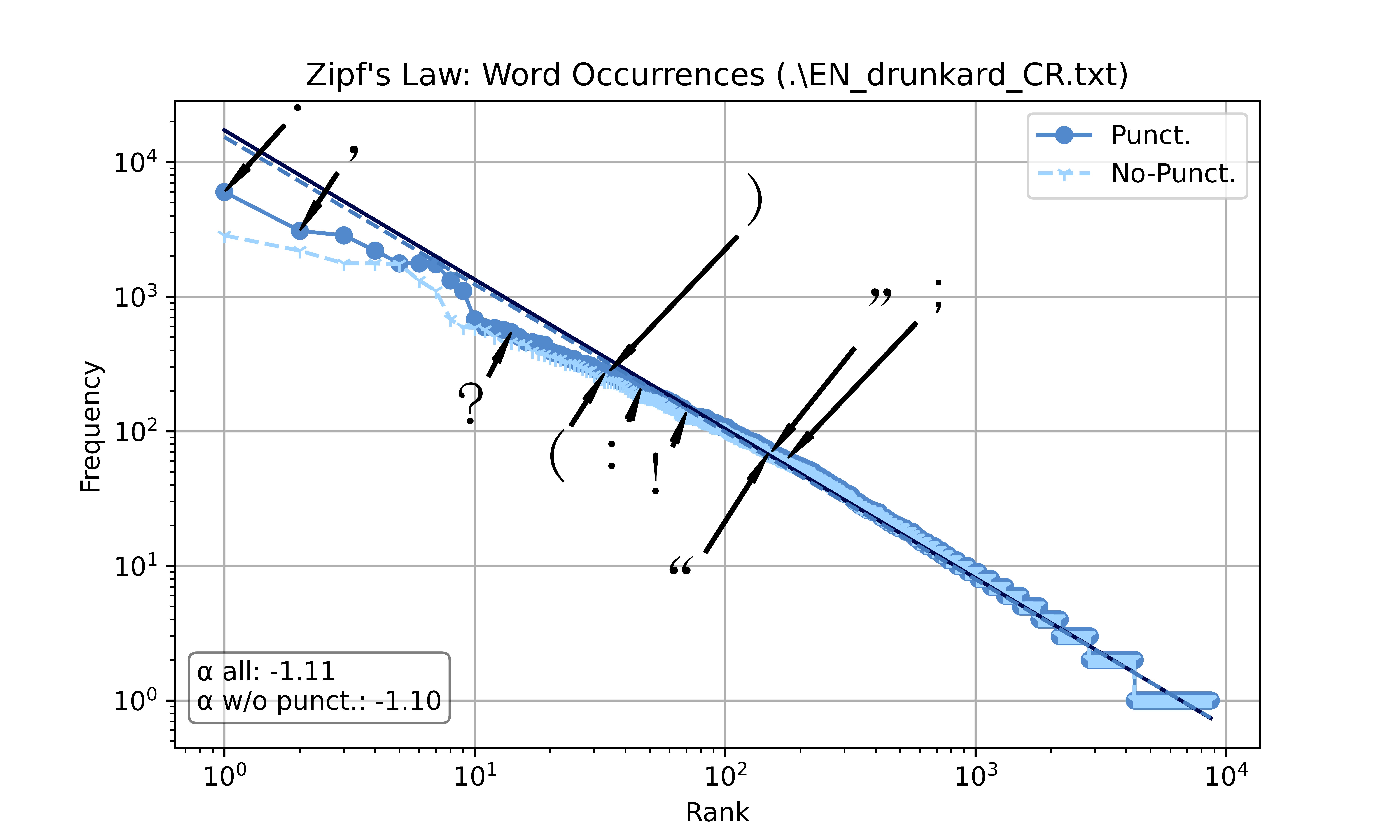}
\includegraphics[width=0.48\textwidth]{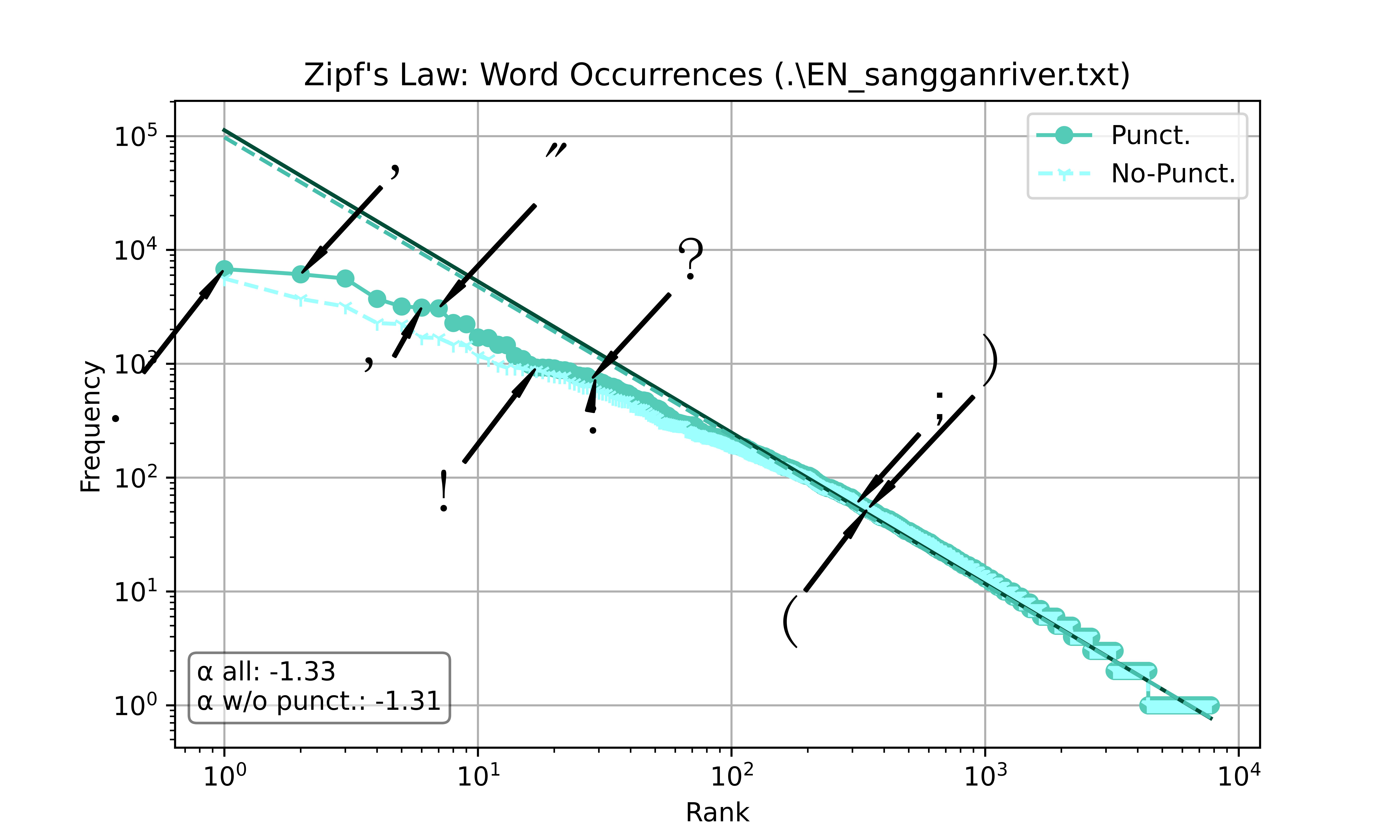}
\includegraphics[width=0.48\textwidth]{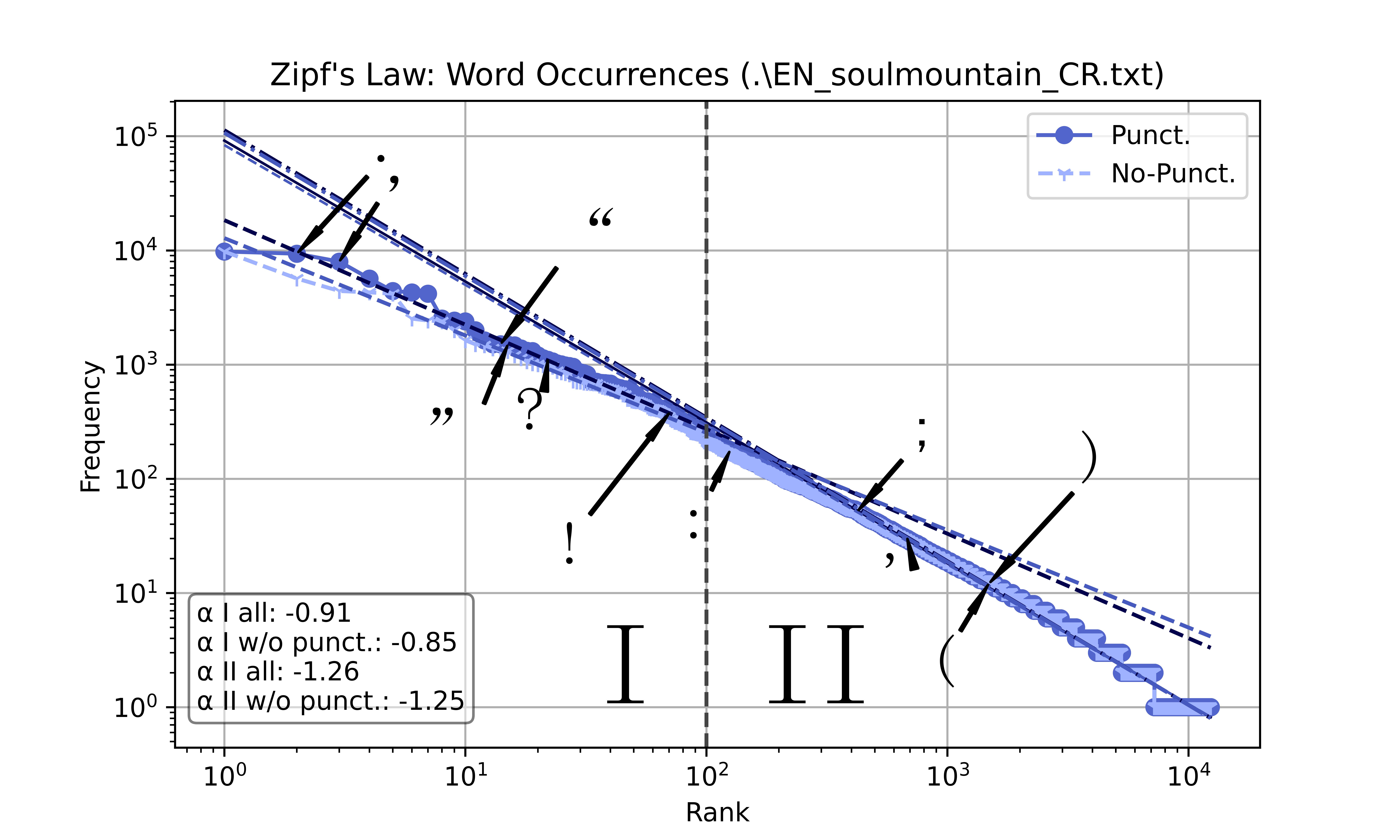}
\caption{The same analysis as that in Fig.~\ref{fig::Zipf_CN} but here for the corresponding English translations.}
\label{fig::Zipf_EN}
\end{figure*}

\subsection{Discrete Weibull distribution}

With regard to the presence of punctuation, writing a text can be considered a process in which, after each consecutive word is written, the writer decides whether to place a punctuation mark or not. To analyze the distribution of distances between consecutive punctuation marks, one can assume that these decisions are random and can be treated as consecutive Bernoulli trials. Using related terminology, placing a punctuation mark can be considered a ``success,'' occurring with some probability $p$, while refraining from using a punctuation mark can be considered a ``failure,'' associated with the probability $1-p$. Thus, writing a sequence of text between one punctuation mark and the next corresponds to a series of trials until the first success. Consequently, the probability mass function (PMF) describing the distances between consecutive punctuation marks (measured by the number of words) follows the distribution of the number of trials until the first success.

If consecutive trials are independent and $p$ is constant, then the relevant distribution is geometric distribution. However, in order to be able to describe situations, in which $p$ changes from trial to trial, the geometric distribution needs to be generalized. One of the possible generalizations is the discrete Weibull distribution~\cite{NakagawaT-1975a}, whose PMF is
\begin{equation}
P(k)=(1-p)^{{(k-1)}^{\beta}}-(1-p)^{{k^{\beta}}}
\label{eq::DWeibull_PMF}
\end{equation}
and cumulative mass function (CMF) is
\begin{equation}
F(k)=1-\left( 1-p \right)^{k^\beta}.
\label{eq::DWeibull_CMF}
\end{equation}

When the probability of the first success $p$ depends on the number of already performed trials $k$, this dependence is called hazard function. For the discrete Weibull distribution, the hazard function is given by
\begin{equation}\label{eq::DWeibull_hazard_function}
h(k) = 1-\left(1-p\right)^{k^{\beta} - (k-1)^{\beta}}.
\end{equation}
The parameter $p=h(1)$ with $0 \le p \le 1$ is the probability of obtaining a success in the first trial. The value of the exponent $\beta > 0$ determines shape of the hazard function. For $\beta=1$, $h(k)$ is constant and discrete Weibull distribution reduces to geometric distribution. For $\beta<1$, $h(k)$ is a monotonously decreasing function (approaching 0 as $k \rightarrow \infty$), while for $\beta>1$ it is an monotonously increasing function (approaching 1 as $k \rightarrow \infty$). The further $\beta$ is from 1, the more rapidly $h$ changes with $k$.

If one studies some data set whose PMF or CMF can potentially be fitted by a discrete Weibull distribution, it is recommended to represent both on the so-called Weibull plots or rescaled Weibull plots. The idea is that, instead of Eq.~(\ref{eq::DWeibull_CMF}), its log-transformed version is used, which is linear in $\log k$. Then a plot in the following coordinate system can be created:
\begin{equation}
x = \log k, \quad y = \log\{-\log[1-F(k)]\},
\end{equation}
on which the discrete Weibull distribution PMF lies on a straight line with slope $\beta$ and intercept $\log[-\log(1-p)]$. As the quantities $k$ and $F(k)$ can also be calculated for a given data set, the respective CMF can directly be compared with the theoretical one. To facilitate a comparison between different discrete Weibull distributions, it is instructive to create the rescaled Weibull plots that use rescaled coordinates $[x',y']$, which fit into a square $[0,1]\times[0,1]$ (for the transformation formulas, see Ref.~\cite{StaniszT-2024b} for example).

\subsection{Multifractal detrended fluctuation analysis}

Multifractal Detrended Fluctuation Analysis (MFDFA) is one of the most popular and reliable methods of identification of fractal properties of time series~\cite{KantelhardtJ-2002a,OswiecimkaP-2006a}. Let us denote by $U=\{u\}_{i=1}^T$ a time series of length $T$ whose fractal properties are to be determined (for example, a time series of sentence lengths). In the first step of the procedure, one creates a time series profile $\hat{U}=\{\hat{u}_i\}_{i=1}^T$ by integrating $U$
\begin{equation}
\hat{u}_i = \sum_{j=1}^i u_j, \qquad i=1,...,T
\end{equation}
and partitions it into a set of disjoint intervals of length $s$ going from both ends of $\hat{U}$ -- this gives $2M_s$ such intervals total. In each interval indexed by $\nu$, a fitted polynomial trend $P_k^m(\nu)$ of order $m$ is subtracted from $\hat{U}$, which gives a detrended residual signal $X=\{x_i\}_{i=1}^T$
\begin{equation}
x_{s\nu+k} = \hat{u}_{s\nu+k} - P_{\nu}^m(k),
\end{equation}
where $k=1,...,s$ and $\nu=0,...,2M_s-1$. In the next step, variance of $X$ is obtained for each segment:
\begin{equation}
f^2(s,\nu) = \frac{1}{s} \sum_{k=1}^s \left( x_{s\nu+k} - \langle x \rangle_{\nu} \right)^2,
\end{equation}
where $\langle x \rangle_{\nu}$ stands for mean of a segment $\nu$. A family of the $q$-dependent fluctuation functions $F_q(s)$ defined by
\begin{eqnarray}
F_q(s) = \bigl\{ \frac{1}{2 M_s} \sum_{\nu=0}^{2M_s-1} \left[ f^2(s,\nu) \right]^{q/2} \bigr\}^{1/q}, \quad q \ne 0,\\
F_q(s) = \exp \bigg\{ \frac{1}{2M_s} \sum_{\nu=0}^{2M_s-1} \ln f^2(s,\nu) \bigg\}, \quad q = 0
\end{eqnarray}
shows a power-law dependence on scale $s$
\begin{equation}
F_q(s) \sim s^{h(q)},
\end{equation}
for all values of $q$ if $U$ is fractal. Otherwise, no such dependence is observed. In practice, the fractal character of $U$ is identified if $F_q(s)$ can be approximated by straight lines on a double logarithmic plot. A function $h(q)$ is the generalized Hurst exponent of $U$ that is decreasing in $q$ if $U$ is multifractal or it is constant $h(q)=H$ if $U$ is monofractal~\cite{BarabasiAL-1991a}, where $H \equiv h(2)$ is the regular Hurst exponent of $U$~\cite{HurstHE-1951a}. The shape of $h(q)$ can play a supportive role in determining the variability range of $q$ for which an analysis may be carried out. This is because the absolute value of $q$ cannot be too large in the case of leptokurtic PMFs where the sufficiently large moments become infinite. In practical situations, the finite size effects can mislead one, because the moments may not diverge. Typically, the allowed variability range of $q$ is inferred from shape of PMF for a given data set, but it can also be inferred from $h(q)$. If the overall shape of $h(q)$ indicates a multifractal character of the data, but $h(q)$ quickly saturates for both positive and negative $q$s, this suggests that the signal amplitude may be cut off at some magnitude. Then, the considered range of $q$ has to be restricted such that $h(q)$ still varies.

Another function commonly considered in the fractal analysis is the singularity spectrum $f(\alpha)$ defined as a Legendre transform of $\tau(q)=qh(q)-1$~\cite{HalseyTC-1986a}:
\begin{eqnarray}
f(\alpha) = q[\alpha(q) - h(q)] + 1,\\
\alpha(q) = h(q) + q \frac{dh(q)}{dq}.
\end{eqnarray}
For a multifractal data set, $f(\alpha)$ is a concave curve with a roughly parabolic shape over some range of its argument: $\Delta\alpha=\alpha_{\rm max}-\alpha_{\rm min}$, while for a monofractal data set $f(\alpha)$ becomes a single point located at $\alpha_0=H$. In this context, it is noteworthy that genuine mutlifractality requires the data set under study being long-range power-law correlated. If this is not the case, the data is either monofractal or bifractal (a two-point $f(\alpha)$ whose support is $\{\alpha_1=0,\alpha_2=H\}$)~\cite{NakaoH-2000a}. Finite size effects can produce a spuriously continuous $f(\alpha)$, therefore one has to be extremely cautious while interpreting results if $\Delta\alpha > 0$~\cite{DrozdzS-2009a,KwapienJ-2023a}.

\section{Results and discussion}

\subsection{Zipf's law}

Fig.~\ref{fig::Zipf_CN} shows the rank-frequency distributions for the three books specified above in two variants: (i) a standard Zipfian approach, which considers only words, and (ii) an extended approach that takes punctuation marks into account alongside words. The degree of compliance of the empirical distributions with the theoretical formula is impressive and not worse than for the European languages analyzed in literature. Moreover, the effect introduced by punctuation is extremely interesting and significant. Similarly to the European languages, punctuation treated on an equal footing with words gives a rank-frequency distribution that fits the Zipf distribution over a broader range of ranks. In particular, the punctuation largely reduces the bias towards lower frequencies at the initial ranks (the most frequent words) which, according to Mandelbrot's proposal, was modeled by adding some constant to rank $R$~\cite{MandelbrotBB-1954a}. To be absolutely precise, the \textit{The Soul Mountain} departs somewhat from the uniform Zipfian behavior, unlike the standard cases represented here by \textit{The Drunkard} and \textit{The Sun Shines over the Sanggan River}. In the former case, one observes two scaling regimes with a crossover at approximately $R=100$: for small ranks, $\gamma \approx 1.19$ (or 1.10 without punctuation marks), while for large ones, $\gamma \approx 0.94$. This decrease of $\gamma$ for $R>100$ indicates a deflection toward a richer vocabulary. It is interesting to note in this connection that a similar rank-frequency distribution occurs in \textit{Finnegans Wake}~\cite{KwapienJ-2012a}, a famous work by James Joyce, which is commonly classified as belonging to the literary style known as 'stream of consciousness'.

The analogous statistics for the available English translations of the considered books is shown in Fig.~\ref{fig::Zipf_EN}. Here, the compliance with Zipf's law is slightly worse than in the original Chinese works. In particular, for large ranks, the index $\gamma$ is significantly larger, which means a faster decrease of the distribution. One possibility is that this may be due to the imperfect reproduction of the original Chinese texts in their translated English equivalents. The faster decline of the distributions in their tails may indicate a reduction in the vocabulary used in the translation.

\subsection{Distribution of the punctuation-mark intervals}

\begin{figure*}
\centering
\includegraphics[width=1.1\textwidth]{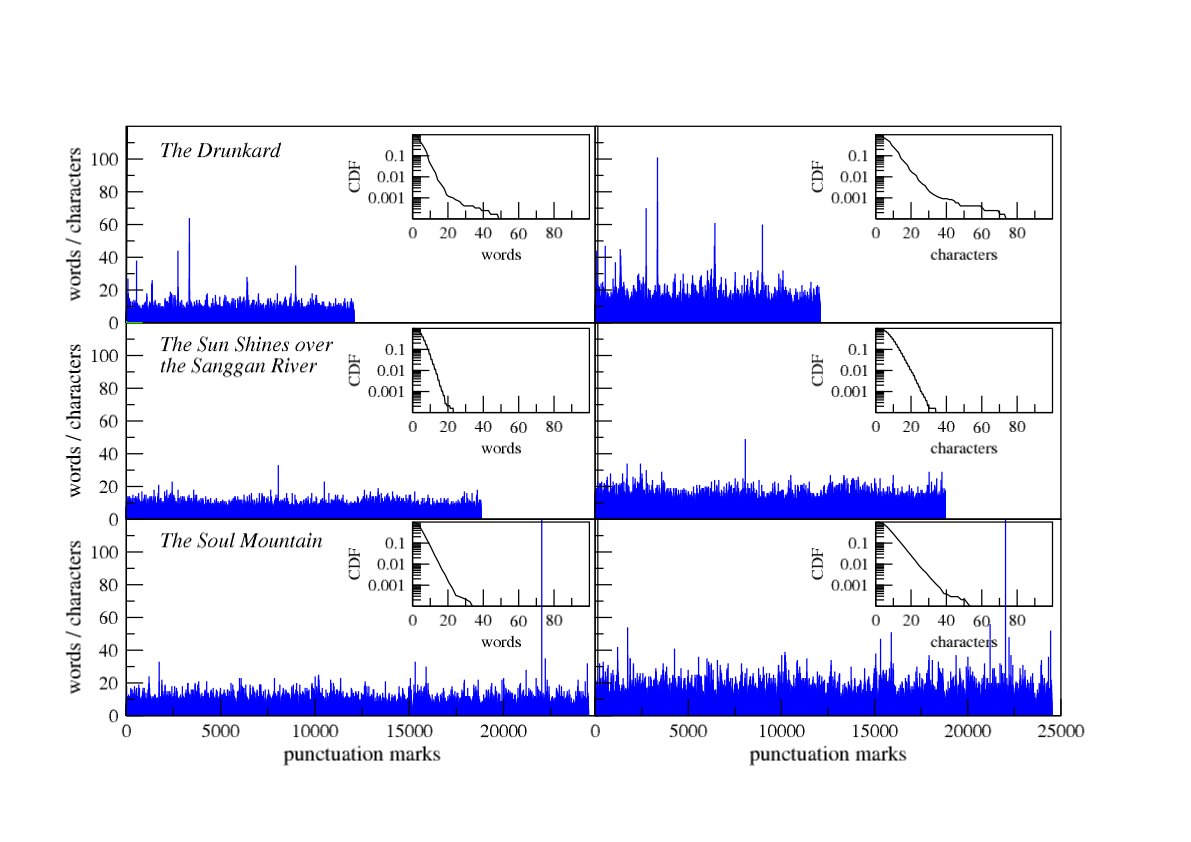}
\caption{Time series of the distances between consecutive punctuation marks measured in words (left) and characters (right) for three Chinese novels: \textit{The Drunkard} (top), \textit{The Sun Shines over the Sanggan River} (middle), and \textit{The Soul Mountain} (bottom). Insets show the cumulative distribution functions for the time series shown in the main panels. The largest data point, at 22,028 in \textit{The Soul Mountain} (bottom), which equals 311 words or 494 characters, has been cut off in order to zoom in on the range of the vertical axis.}
\label{fig::time.series.punctuation.mark.distances}
\end{figure*}

Distance between a pair of consecutive punctuation marks can be expressed by (i) the number of characters and (ii) the number of words. For the Chinese novels considered here, the fluctuations of this quantity are shown in Fig.~\ref{fig::time.series.punctuation.mark.distances}. Of course, the distance measured by the number of symbols (left column)) is larger, on average, by a factor of about 1.5 (in Chinese, some symbols themselves constitute words) than the distance measured by the number of words (right column). However, the relative magnitude of distance variability is similar for both ways of measurement. One can also observe that the fluctuations of these intervals are more pronounced in the case of \textit{The Drunkard} and \textit{The Soul Mountain} than in the case of \textit{The Sun Shines Over the Sanggan River}. This finds confirmation in the distributions of these quantities shown in the insets. Clearly, the distributions corresponding to the actual novels develop thicker tails than the Gaussian ones but \textit{The Sun Shines Over the Sanggan River} is least pronounced in this regard. It is also interesting to observe that, relative to the bulk, these distributions decay more slowly if the distance is measured by the number of words than by the number of characters.

\begin{figure*}
\centering
\includegraphics[width=1.1\textwidth]{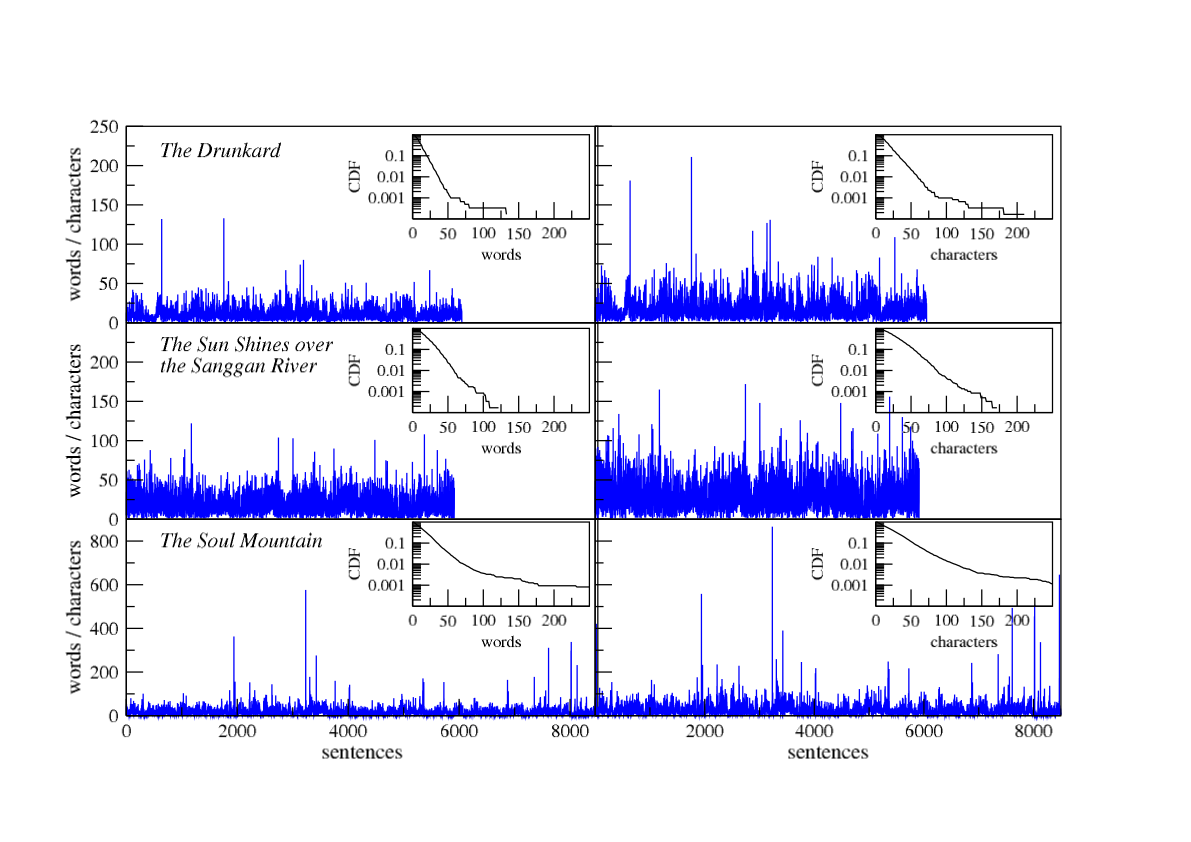}
\caption{The same quantities as in Fig.~\ref{fig::time.series.punctuation.mark.distances} but here for the consecutive sentence lengths measured in words (left) and characters (right). Note that the vertical axis in the bottom panels differs from the one in the top and middle panels.}
\label{fig::time.series.sentence.length}
\end{figure*}

\begin{figure*}
\centering
\includegraphics[width=1.1\textwidth]{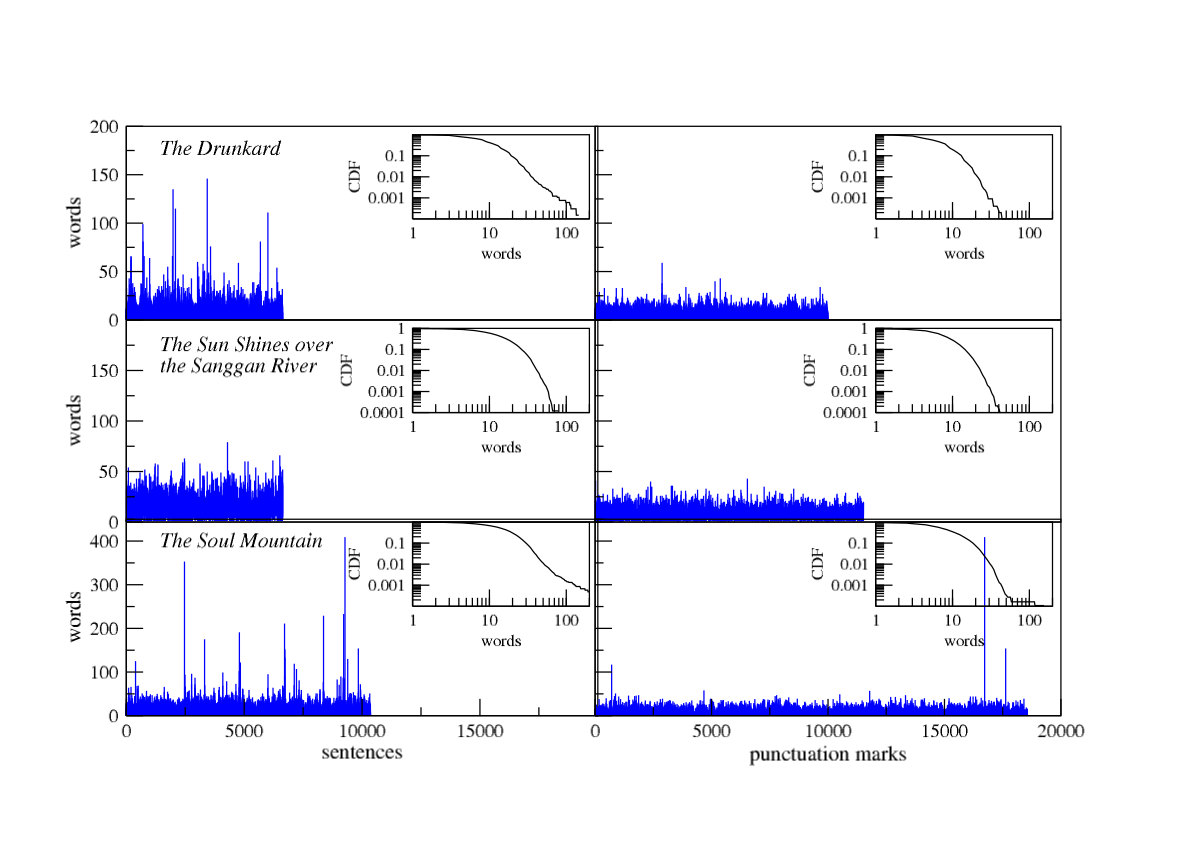}
\caption{Time series of sentence lengths (left) and time series of the distances between consecutive punctuation marks (right) for the English translations of the same Chinese novels as in Figs.~\ref{fig::time.series.punctuation.mark.distances} and~\ref{fig::time.series.sentence.length}. Please note the different scales of the vertical axis. Insets show the cumulative distribution functions for the time series shown in the main panels.}
\label{fig::time.series.punctuation.mark.distances.en}
\end{figure*}

A special subset of punctuation marks is those indicating the ends of sentences. It includes the period, of course, but also the exclamation mark, question mark, and ellipsis. The distance between successive marks indicates the length of the related sentence. Time series representing the consecutive sentence lengths are displayed in Fig.~\ref{fig::time.series.sentence.length} for distance measured in words (left column) and characters (right column). The numbers are now significantly larger than in Fig.~\ref{fig::time.series.punctuation.mark.distances}, which is obvious, but the proportions in their variability are also noticeably larger, especially in the case of \textit{The Soul Mountain} where the variability of sentence lengths has a distinctly cascading nature. The distribution of the corresponding sentence lengths, as shown in the insets, develops tails that are thicker than in Fig.~\ref{fig::time.series.punctuation.mark.distances}, consistently with this observation.

It is interesting to juxtapose these results obtained for the original Chinese texts with their English translations. Since English uses the alphabetic writing system, no significant qualitative difference is expected to exist between the quantities of interest measured in characters and words. This is why only the distances measured in words are considered here. The respective time series of sentence lengths and distances between consecutive punctuation marks together with their PMFs are collected in Fig.~\ref{fig::time.series.punctuation.mark.distances.en}. In general, the characteristics of the sentence-length fluctuations in English do not differ much from their counterparts in Chinese. Only in the case of \textit{The Drunkard} there are slightly more long sentences (length > 50) in the translated text than in the original one (compare with Fig.~\ref{fig::time.series.sentence.length}). Sentence lengths in \textit{The Sun Shines over the Sanggan River} develop CMF with the thinnest tail, while CMF of the same quantity in \textit{The Soul Mountain} -- the thickest ones. This remains in agreement with what was observed for the Chinese texts. A visible difference regarding the distances between consecutive punctuation marks is that the English punctuation in the three translated texts suppresses the largest distances that were seen in the Chinese sources in Fig.~\ref{fig::time.series.punctuation.mark.distances}. Due to the fact that the text sample is too small to draw decisive conclusions on the origin of this difference, it can be noted that they can be related to the differences in grammar between both languages or to some specific properties of the considered translations.

\begin{figure*}
\centering
\includegraphics[trim={0cm 0 0 0.36cm}, clip, width=0.4\textwidth]{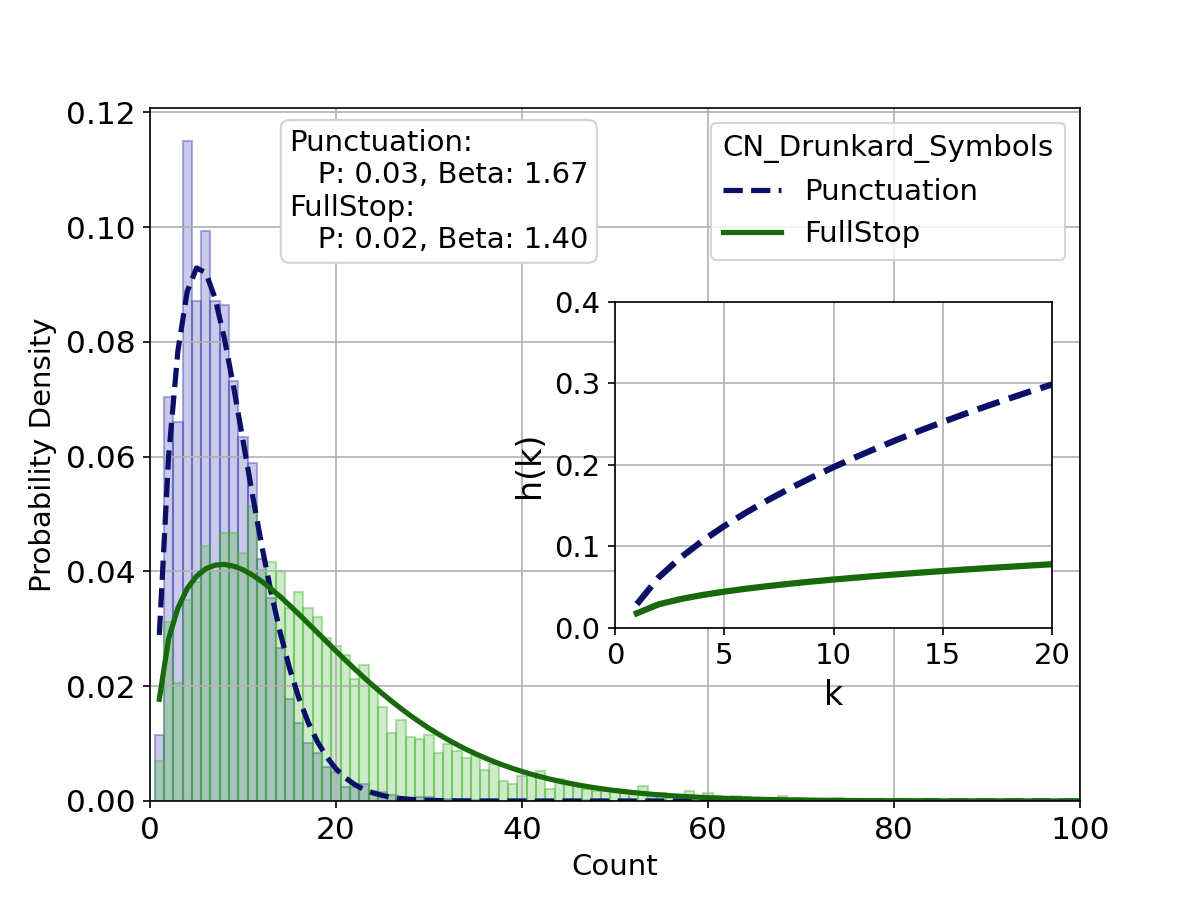}
\includegraphics[trim={0cm 0 0 0.36cm}, clip, width=0.4\textwidth]{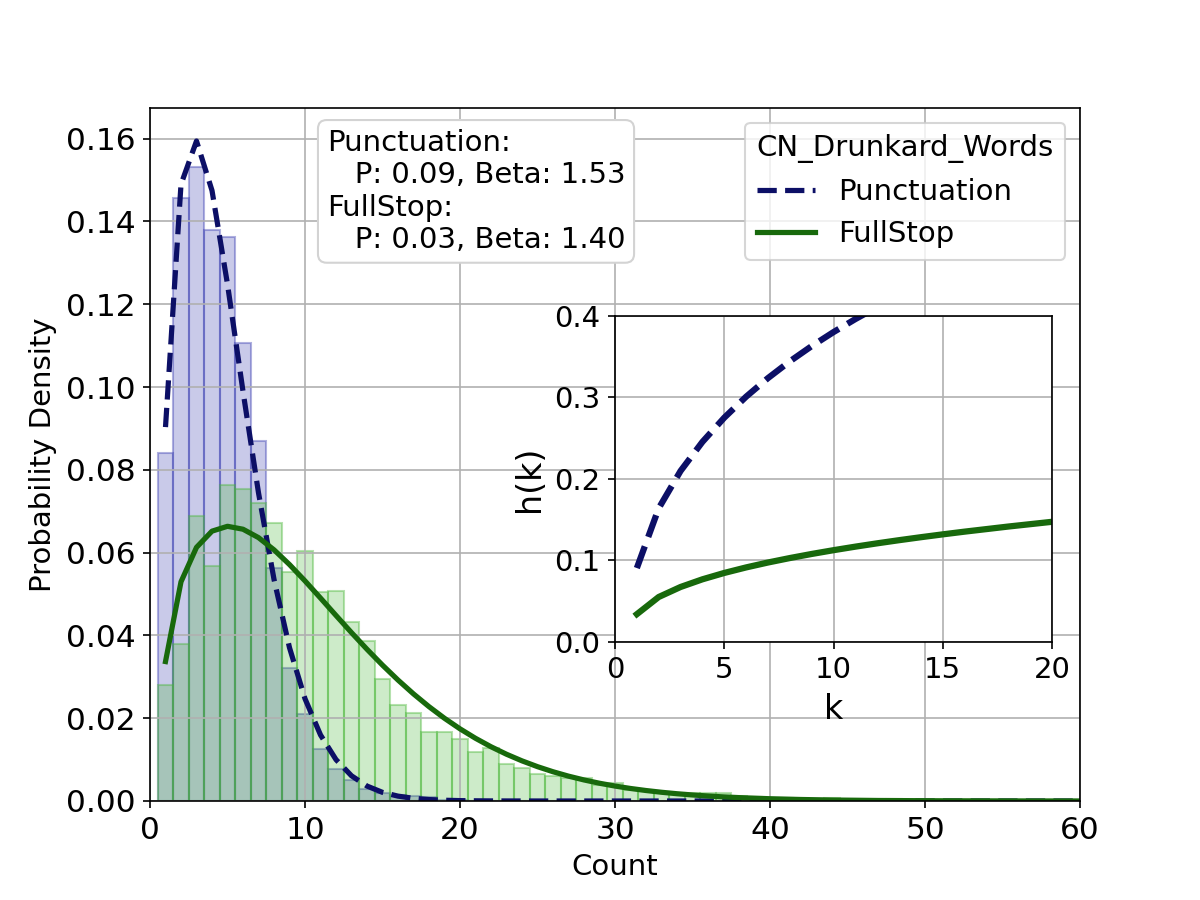}
\includegraphics[trim={0cm 0 0 0.36cm}, clip, width=0.4\textwidth]{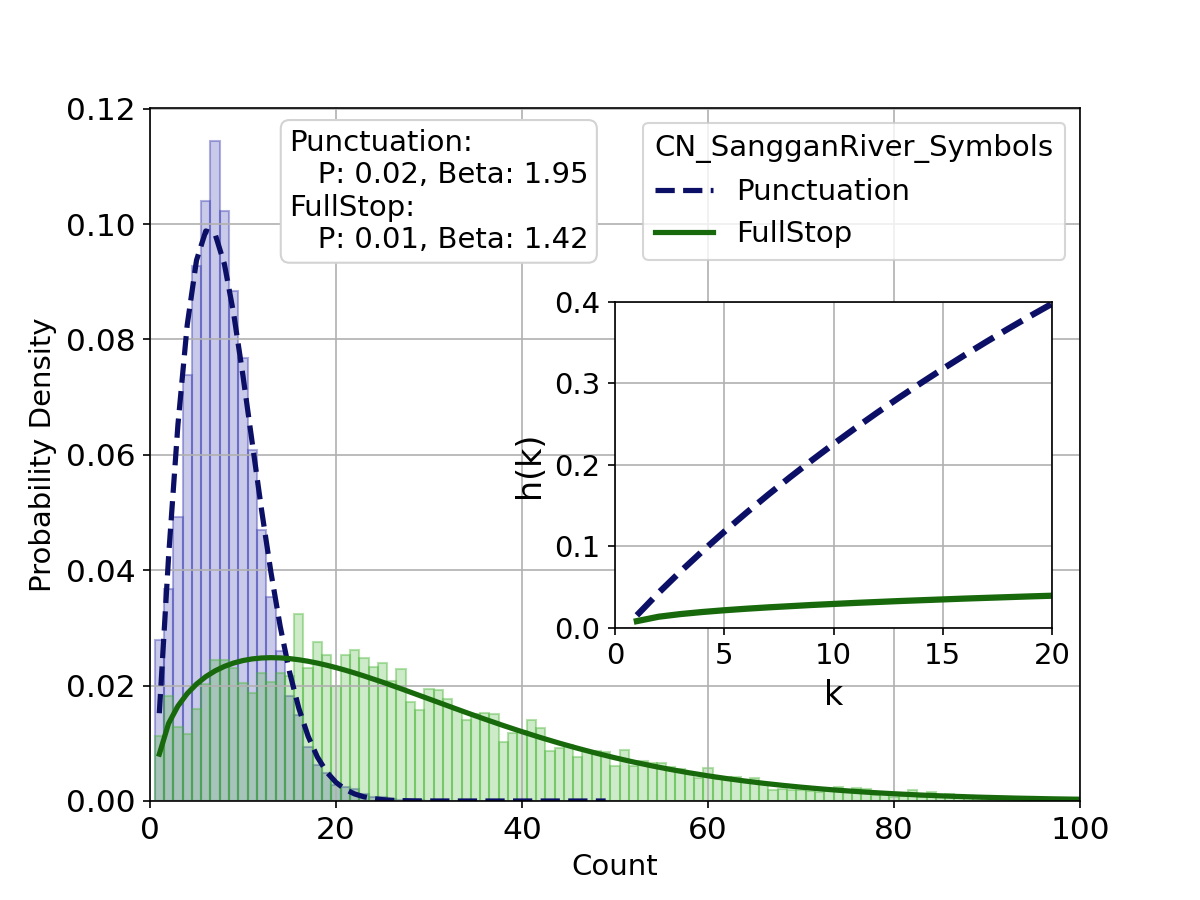}
\includegraphics[trim={0cm 0 0 0.36cm}, clip, width=0.4\textwidth]{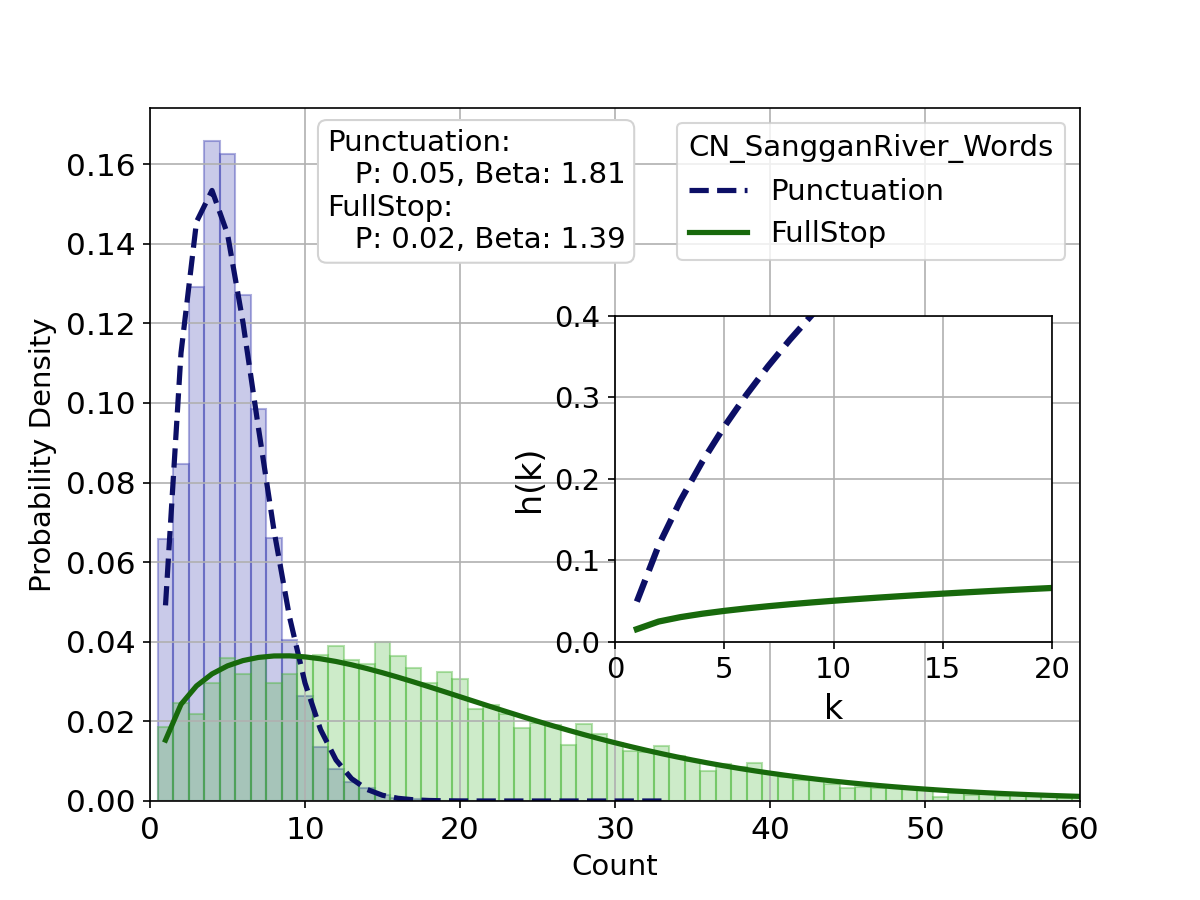}
\includegraphics[trim={0cm 0 0 0.36cm}, clip, width=0.4\textwidth]{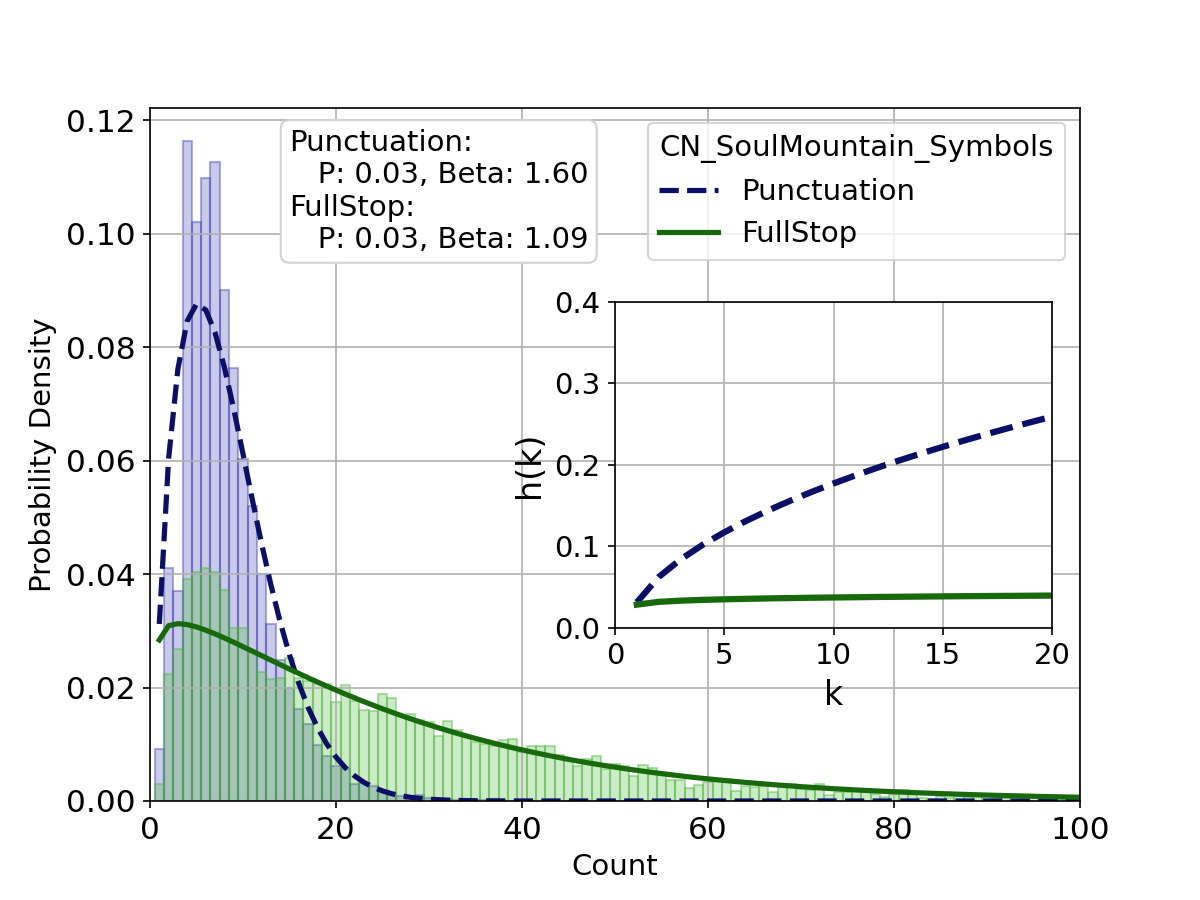}
\includegraphics[trim={0cm 0 0 0.36cm}, clip, width=0.4\textwidth]{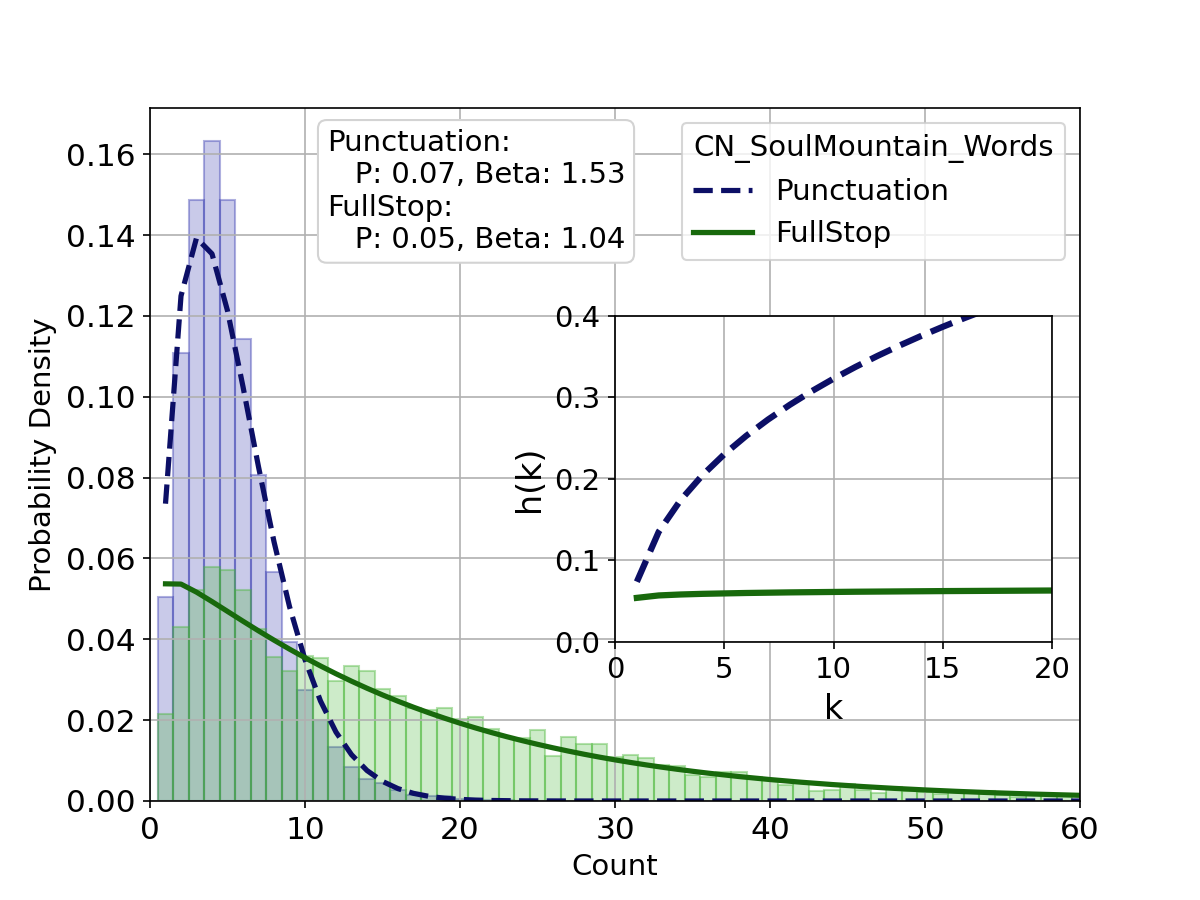}
\caption{The empirical probability mass functions of sentence lengths (green histograms) and distances between consecutive punctuation marks (blue histograms) measured in characters (left) and words (right) for three Chinese novels: \textit{The Drunkard} (top), \textit{The Sun Shines over Sanggan River} (middle), and \textit{The Soul Mountain} (bottom). The discrete Weibull distributions best-fitted to each histogram are denoted by dashed lines (sentence lengths) and solid lines (distances between punctuation marks). In each panel, the fitted values of $p$ and $\beta$ of the discrete Weibull distribution are provided (see Eq.~(\ref{eq::DWeibull_CMF})). Insets show the corresponding hazard functions $h(k)$ defined by Eq.(\ref{eq::DWeibull_hazard_function}).}
\label{fig::Weibull_CN}
\end{figure*}

The full quantitative representation of these characteristics in terms of the discrete Weibull distribution (Eq.~(\ref{eq::DWeibull_CMF})) is displayed in the main panels of Fig.~\ref{fig::Weibull_CN}. The plots show that the fitted distributions roughly agree with the empirical data in all the considered cases. However, the fit quality seems to be better systematically if the distances are measured in words than characters. Also, in either case, the fit quality is better if all the punctuation marks are included rather than only the sentence-ending marks are considered. This, in fact, parallels the observations made earlier~\cite{StaniszT-2023a,StaniszT-2024a} for the Western languages. For the distances between punctuation marks, the specific values of the parameter $\beta$ in the fitted discrete Weibull distributions tend to be larger in the Chinese books considered in this study ($1.5 \le \beta \le 1.9$) than those reported by the previous studies based on English ($0.9 \le \beta \le 1.6$), Russian ($1.1 \le \beta \le 1.5$), and Spanish ($1.0 \le \beta \le 1.4$) texts, while the values of $\beta$ for the Chinese books seem to overlap more with those for German ($1.2 \le \beta \le 1.8$), Italian ($1.1 \le \beta \le 1.7$), Polish ($1.1 \le \beta \le 1.7$), and French ($1.0 \le \beta \le 1.8$) texts~\cite{StaniszT-2023a,StaniszT-2024a,StaniszT-2024b}. Such relatively large values indicate that there is a tendency for the distribution tails to become thinner for the Chinese texts considered here (since the limiting value of $\beta \to 2$ corresponds to a Gaussian). The situation looks different for the sentence lengths in \textit{The Drunkard} and \textit{The Soul Mountain} ($1.4 \le \beta \le 1.5$), but not for \textit{The Sun Shines over the Sanggan River} which shows $\beta$ matching its values for English books ($\beta \approx 1.0$). A more systematic study of the Chinese literature is needed to establish a more decisive evidence, though.

\begin{figure*}
\centering
\subfloat[]{\includegraphics[width=0.4\textwidth]{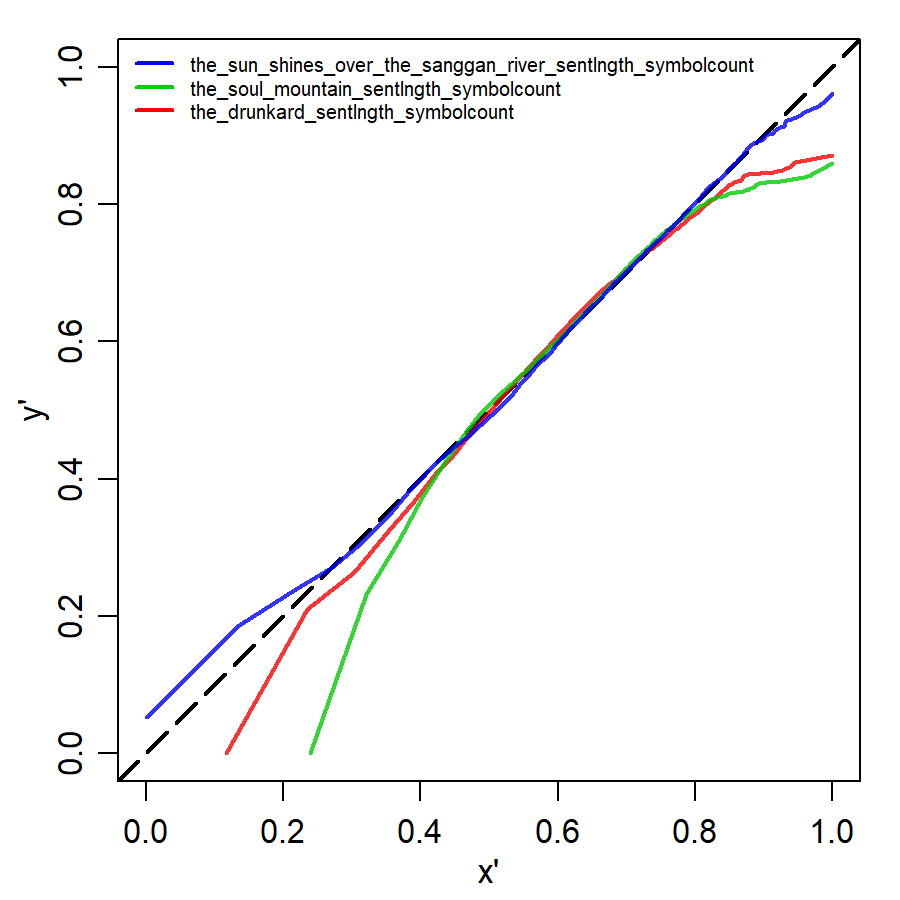}}
\qquad
\subfloat[]{\includegraphics[width=0.4\textwidth]{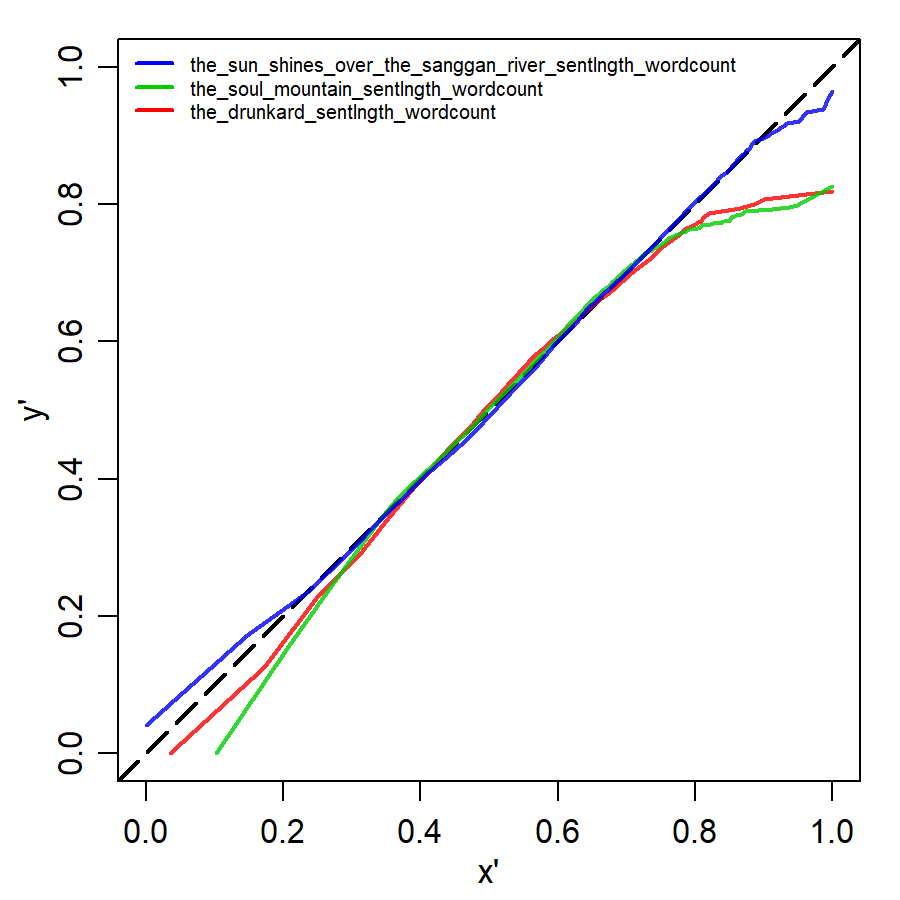}}

\subfloat[]{\includegraphics[width=0.4\textwidth]{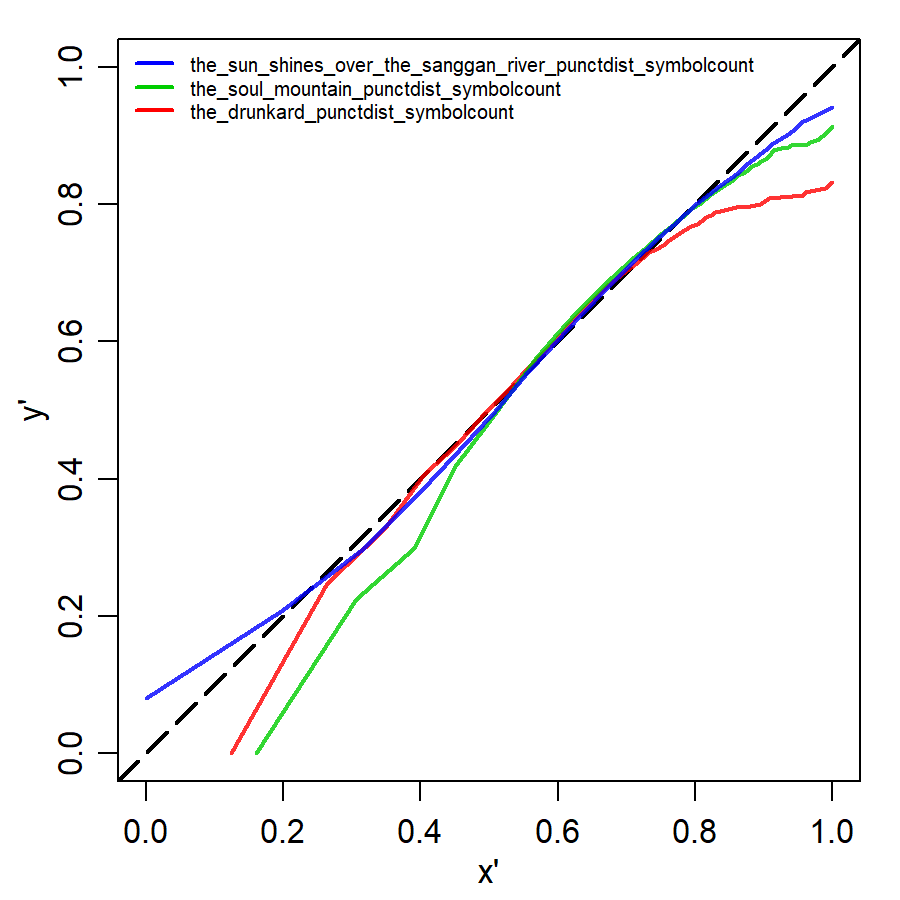}}
\qquad
\subfloat[]{\includegraphics[width=0.4\textwidth]{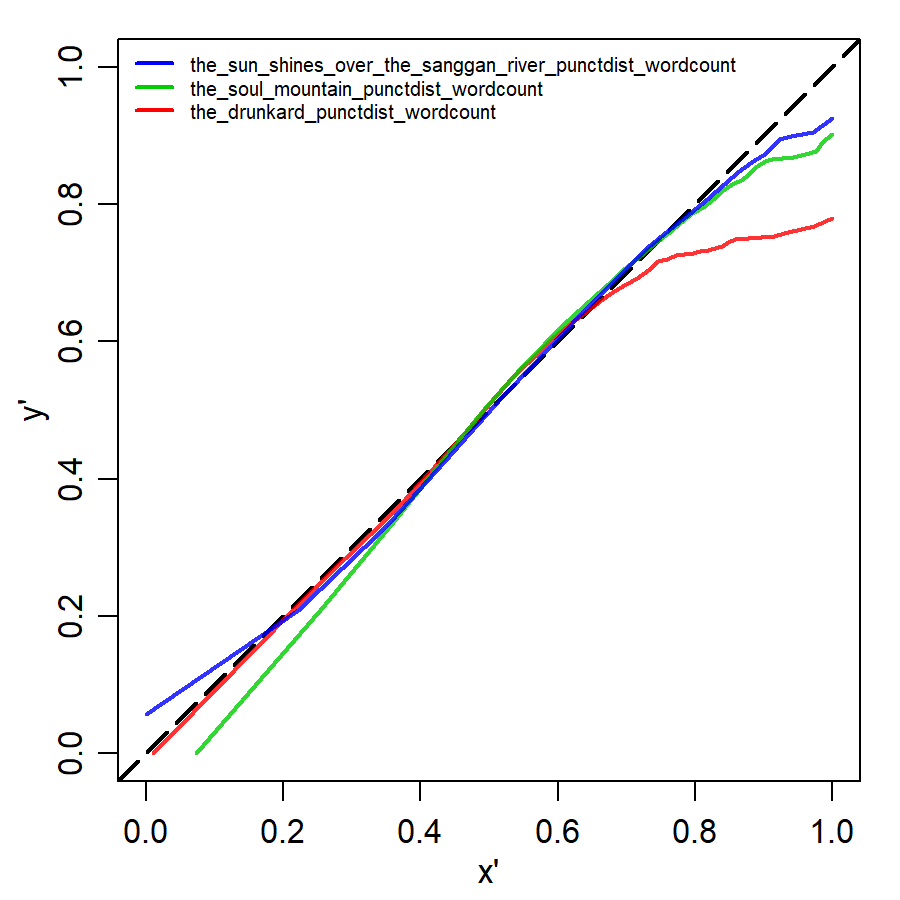}}
\caption{The rescaled Weibull plots for time series of sentence lengths measured in characters (a), sentence lengths measured in words (b), distances between consecutive punctuation marks measured in characters (c), and distances between consecutive punctuation marks measured in words (d). The results for three Chinese novels are shown in each panel: \textit{The Drunkard} (red dashed), \textit{The Sun Shines over the Sanggan River} (blue dash-dotted), and \textit{The Soul Mountain} (green solid).}
\label{fig::rescaled_Weibull_plots}
\end{figure*}

To what extent the empirical data matches the discrete Weibull distributions, one can infer from Fig.~\ref{fig::rescaled_Weibull_plots} that shows the rescaled Weibull plots for all the considered time series. It occurs that the closest agreement is observed for \textit{The Sun Shinesover the Sanggan River}, whose plot is almost linear for the broadest range of argument $x'$ independent of whether the sentence lengths or punctuation-mark distances are considered and independent of whether these quantities are measured in characters or words. The plots representing both \textit{The Soul Mountain} and \textit{The Drunkard} deflect much more from the theoretical form than the plot for Ding Ling's book. This agrees, of course, with the results shown in Fig.~\ref{fig::Weibull_CN} but the rescaled plots in Fig.~\ref{fig::rescaled_Weibull_plots} allow for drawing conclusions more easily.

\begin{figure*}
\centering
\includegraphics[trim={0cm 0 0 0.36cm}, clip, width=0.4\textwidth]{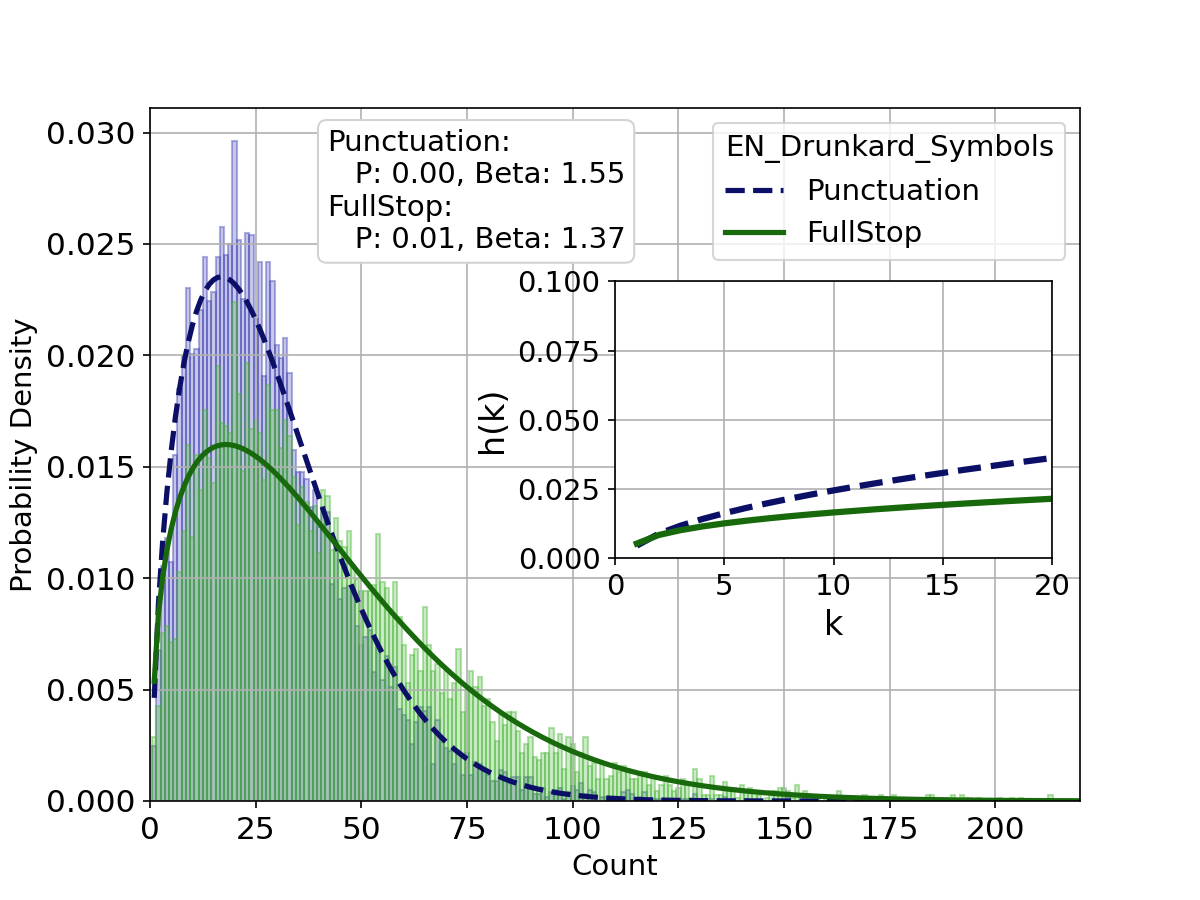}
\includegraphics[trim={0cm 0 0 0.36cm}, clip, width=0.4\textwidth]{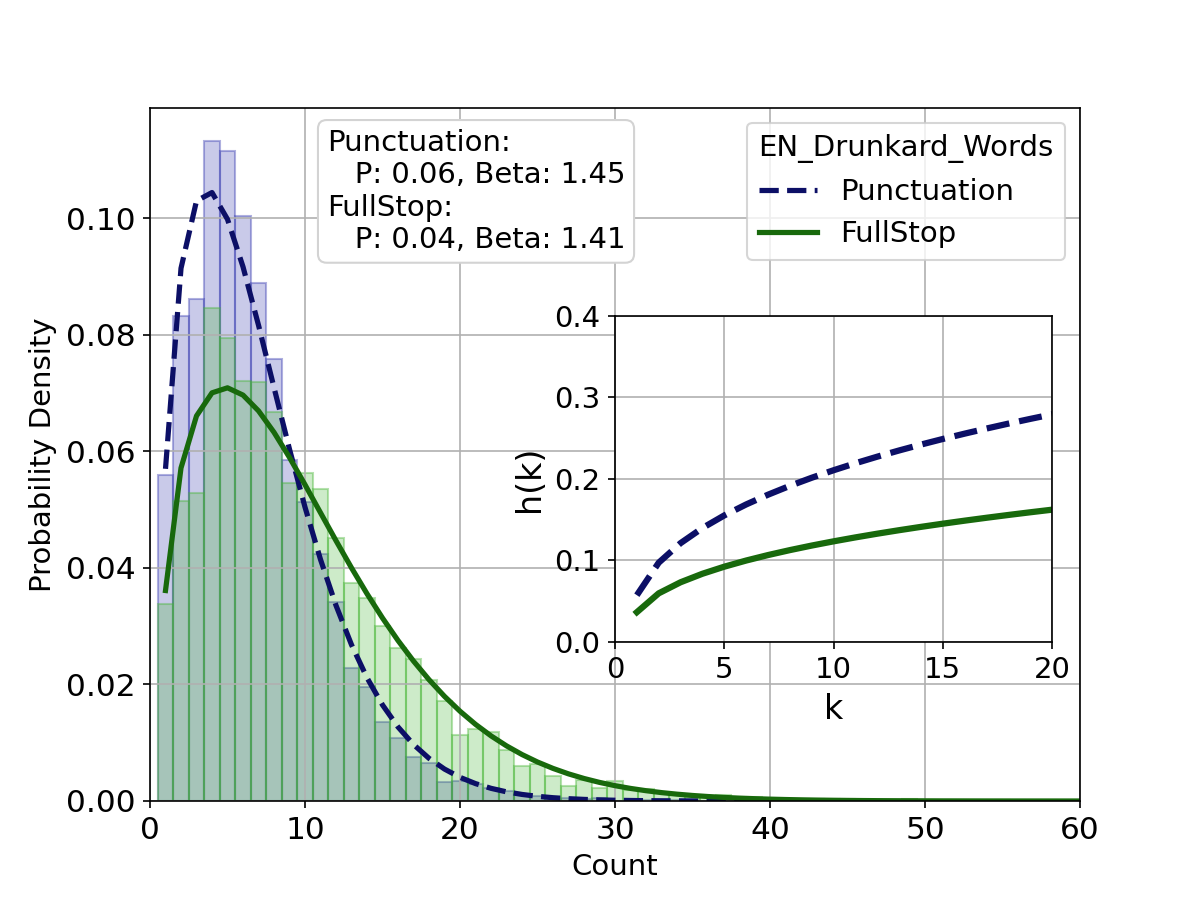}
\includegraphics[trim={0cm 0 0 0.36cm}, clip, width=0.4\textwidth]{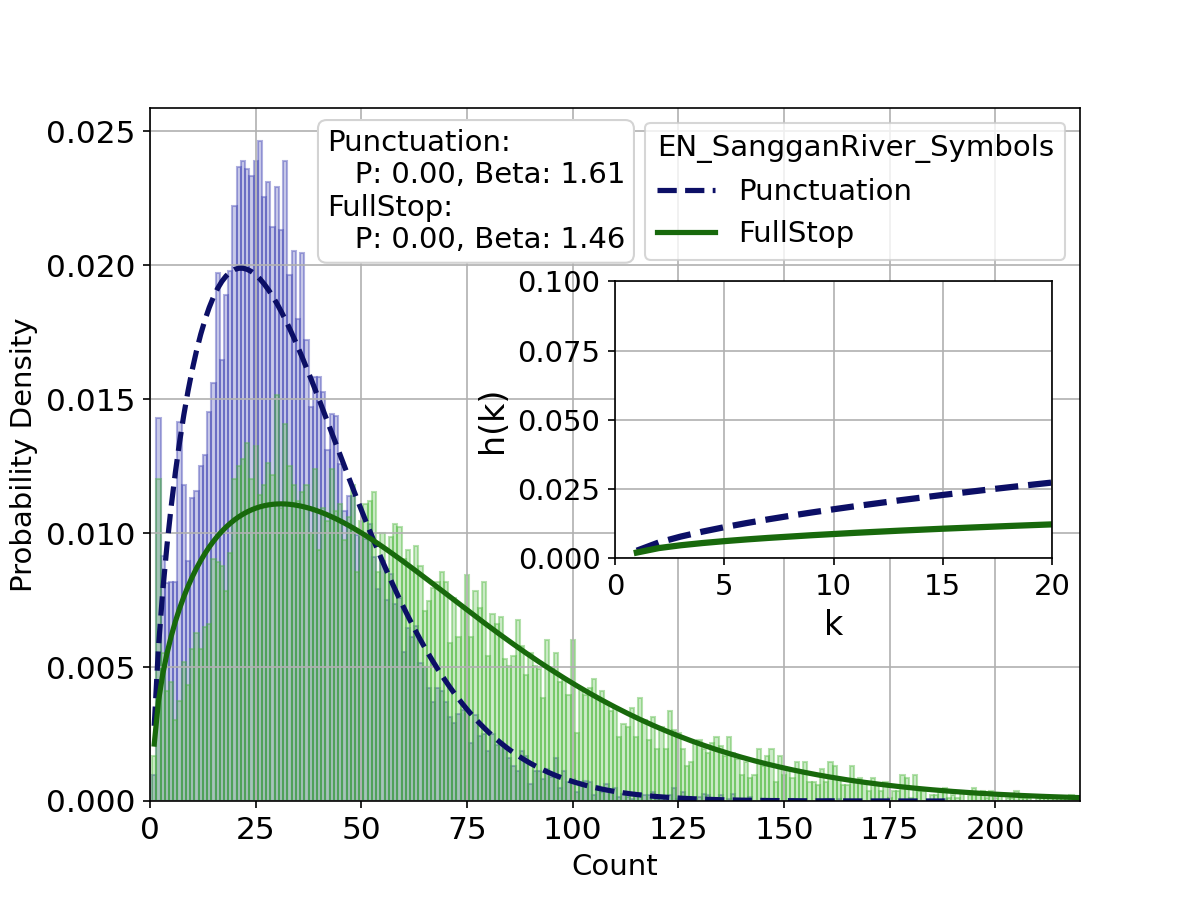}
\includegraphics[trim={0cm 0 0 0.36cm}, clip, width=0.4\textwidth]{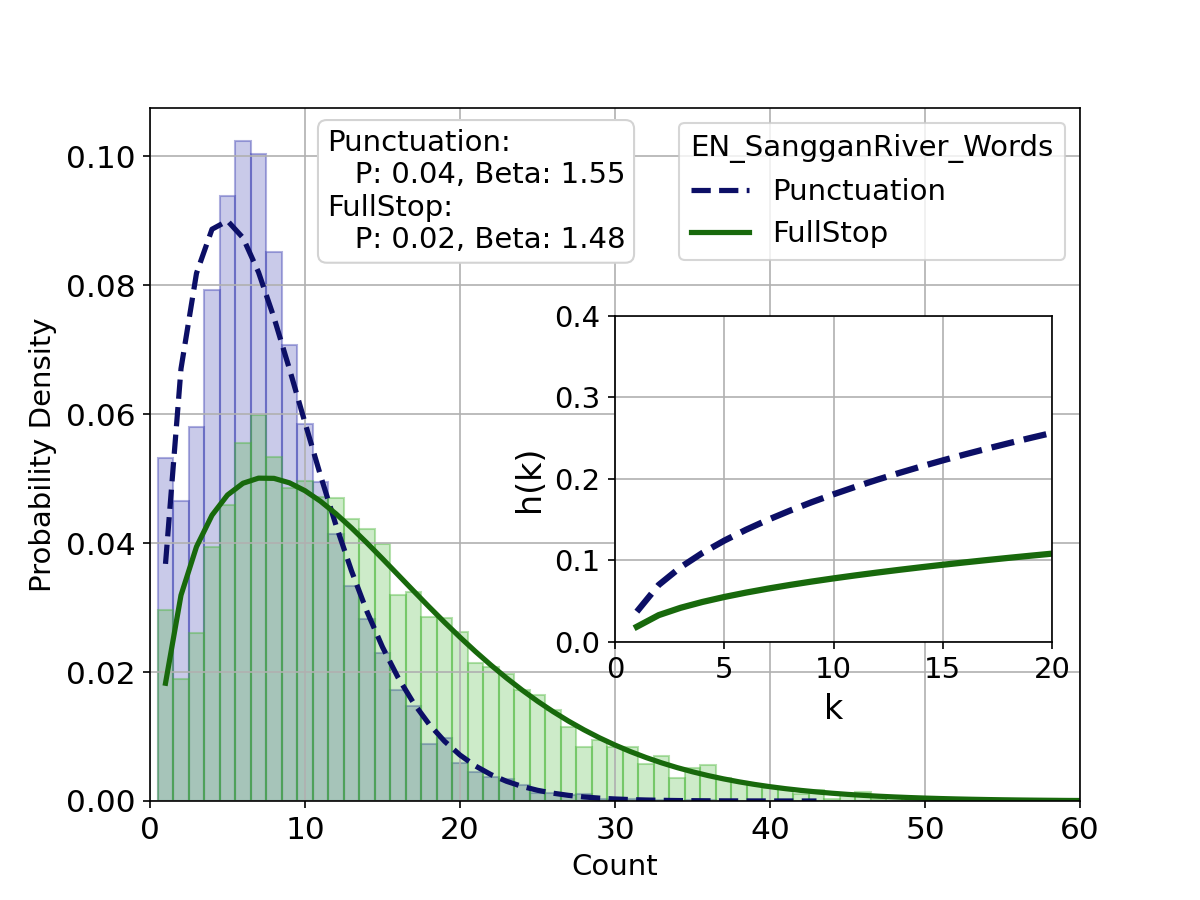}
\includegraphics[trim={0cm 0 0 0.36cm}, clip, width=0.4\textwidth]{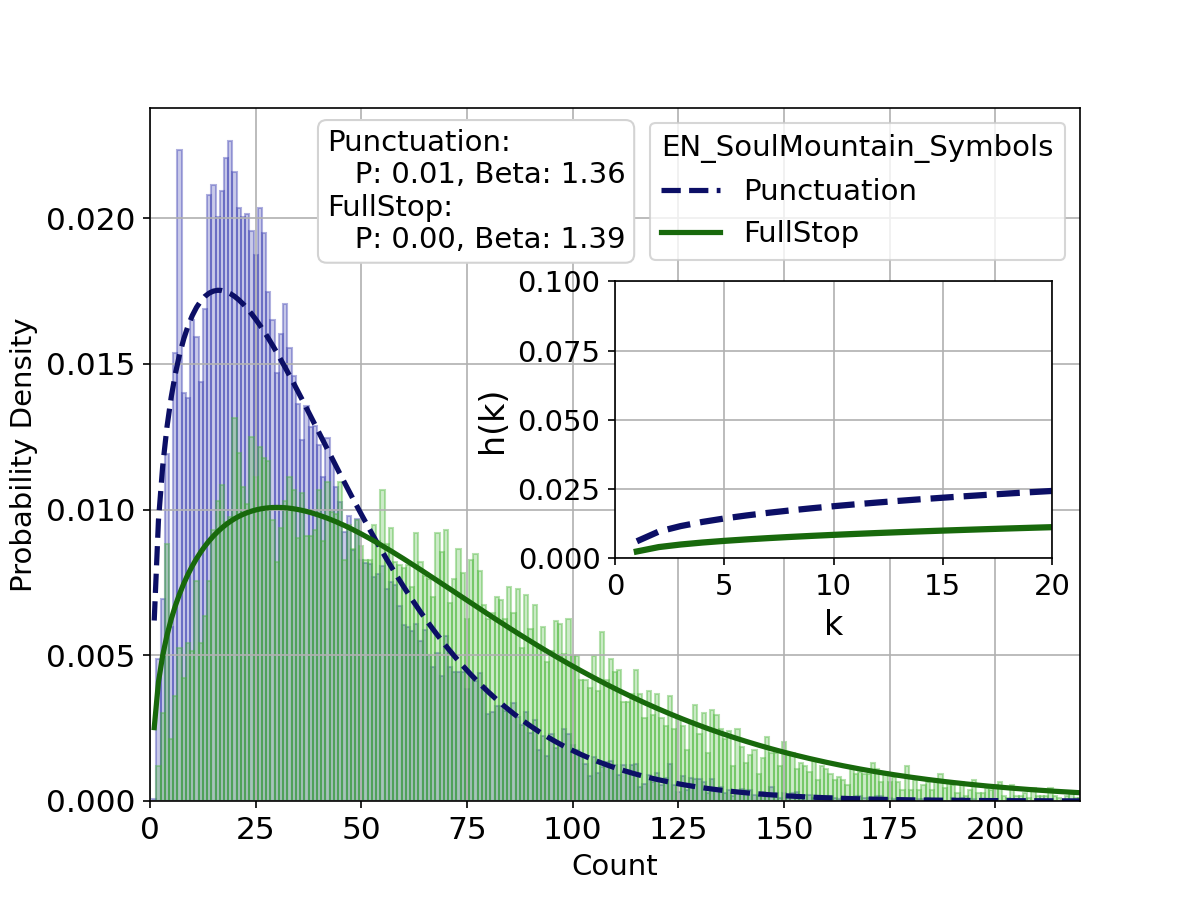}
\includegraphics[trim={0cm 0 0 0.36cm}, clip, width=0.4\textwidth]{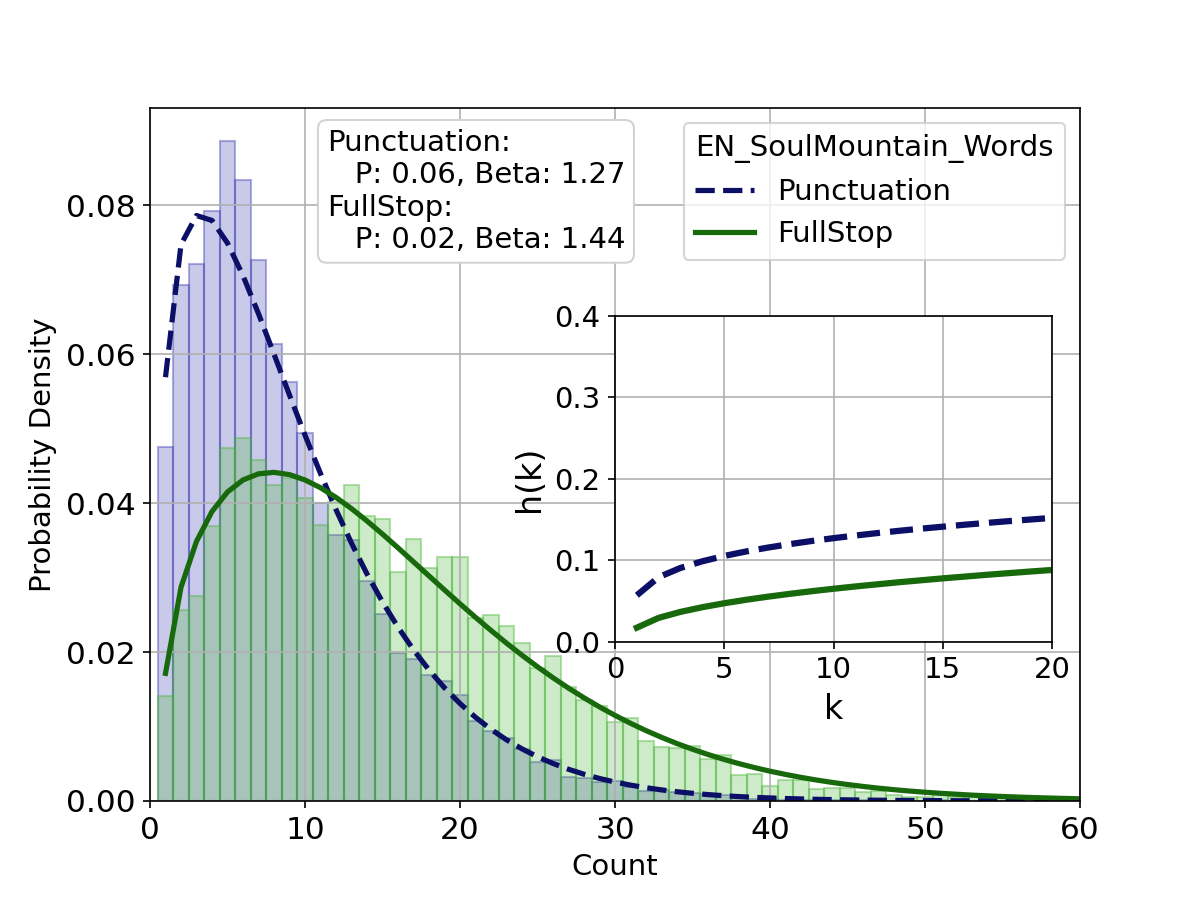}
\caption{The same quantities as in Fig.~\ref{fig::Weibull_CN} but here for the English translations of the respective novels.}
\label{fig::Weibull_EN}
\end{figure*}

The PMFs calculated from the time series representing the English translations of all three books are shown in the main panels of Fig.~\ref{fig::Weibull_EN} as histograms. In each case, the histograms for both the sentence lengths and the inter-mark distances are shown on the same panel. The empirical distributions are accompanied by the best-fitted discrete Weibull distributions with their parameters $p$ and $\beta$ given explicitly. If the inter-mark distances are considered, the English translations show smaller values of $\beta$ than their Chinese sources, which confirms the observation made from Fig.~\ref{fig::Weibull_CN}. This is not the case for the sentence lengths, where $\beta \approx 1.4$ is either invariant under translation CN$\to$EN or increases from $1.0$ for Chinese to this value for English (bottom panels of Figs.~\ref{fig::Weibull_CN} and~\ref{fig::Weibull_EN}).

These characteristics are even more clearly visible in the representation of the hazard function $h(k)$ defined by Eq.(\ref{eq::DWeibull_hazard_function}) and presented in the insets of Figs.~\ref{fig::Weibull_CN} and~\ref{fig::Weibull_EN}. $h(k)$ is monotonously increasing since $\beta > 1$ in each case. For the distances between characters marking the ends of sentences, the functions $h(k)$ are rather comparable for the original Chinese texts and their English translations. However, when all punctuation is taken into account, the function $h(k)$ increases significantly stronger with $k$ for a Chinese text than for its English counterpart regardless of which book is analyzed. This is an interesting result, which indicates a much more frequent need to use punctuation in Chinese texts.

\subsection{Multifractal analysis}

\begin{figure*}
\centering
\includegraphics[width=1.0\textwidth]{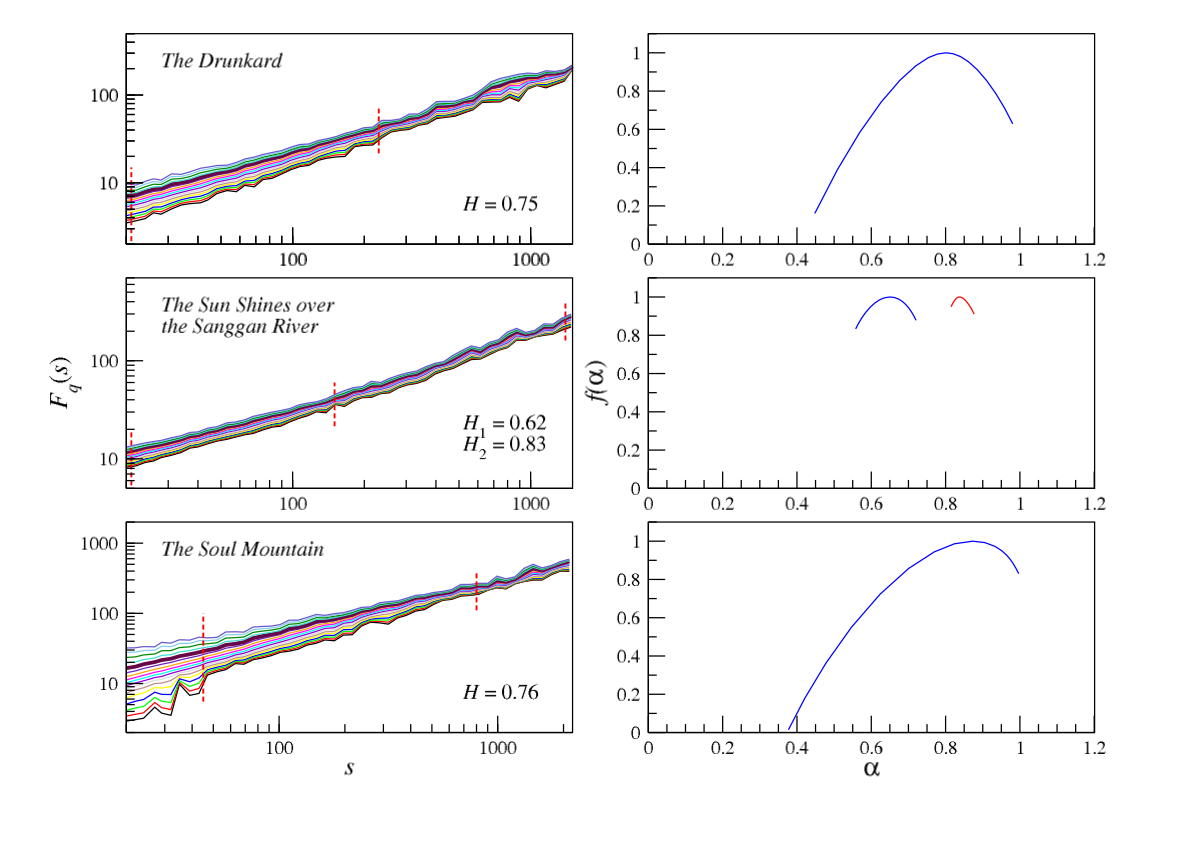}
\caption{A family of the fluctuation functions $F_q(s)$ with $-4 \le q \le 4$ (left) and the corresponding singularity spectra $f(\alpha)$ (right) calculated for time series of sentence lengths measured in words for three Chinese novels: \textit{The Drunkard} (top), \textit{The Sun Shines Over Sanggan River} (middle), and \textit{The Soul Mountain} (bottom). In the last case, the largest observation, which can be viewed as an outlier, has been removed from the respective time series. Vertical dashed red lines in each plot on the left show the range of approximately power-law dependence of the fluctuation functions. In the case of \textit{The Sun Shines...} two such ranges have been identified, which results in two singularity spectra shown on the right.}
\label{fig::Fq.sentence.lengths.words}
\end{figure*}

Even a quick glance at the time series presented in Figs.~\ref{fig::time.series.punctuation.mark.distances}-\ref{fig::time.series.punctuation.mark.distances.en}, allows one to infer that there are both fluctuation amplitude clustering (i.e., long-range correlations) and large fluctuations (i.e, a possible cascade-like structure) present. This opens space for a multifractal analysis of the data. To start with, time series of sentence lengths measured in words for each book are selected and subject to MFDFA. Because the respective CMFs develop heavy tails, the range of the variability of $q$ has been restricted to $-4 \le q \le 4$. The resulting $q$-dependent fluctuation functions are shown in Fig.~\ref{fig::Fq.sentence.lengths.words} (left column) as log-log plots. For all three books, it is possible to distinguish such a range of scales $s$ that a firm power-law dependence is observed (\textit{The Drunkard} and \textit{The Soul Mountain}) or even two such ranges of $s$ (\textit{The Sun Shines over the Sanggan River}). It is noteworthy that scaling is evident for the whole considered range of $q$ in each case. Values of the Hurst exponent are close to $H\approx 0.75$ for the books without the scaling-regime crossover and these change from $H=0.62$ to $0.83$ for the one with two scaling regimes. These values are in general agreement with the ones reported in Ref.~\cite{LiuJ-2023a}. \textit{The Drunkard} and \textit{The Soul Mountain} present broad singularity spectra with $\Delta\alpha > 0.5$, which indicates that the related time series are strongly multifractal. On the other hand, the two calculated spectra for \textit{The Sun Shines over the Sanggan River} are much narrower. The long-scale one is rather monofractal ($\Delta\alpha \approx 0.05$), while the short-scale $f(\alpha)$ is either weakly multifractal or indecisive ($\Delta\alpha \approx 0.15$).

\begin{figure*}
\centering
\includegraphics[width=1.0\textwidth]{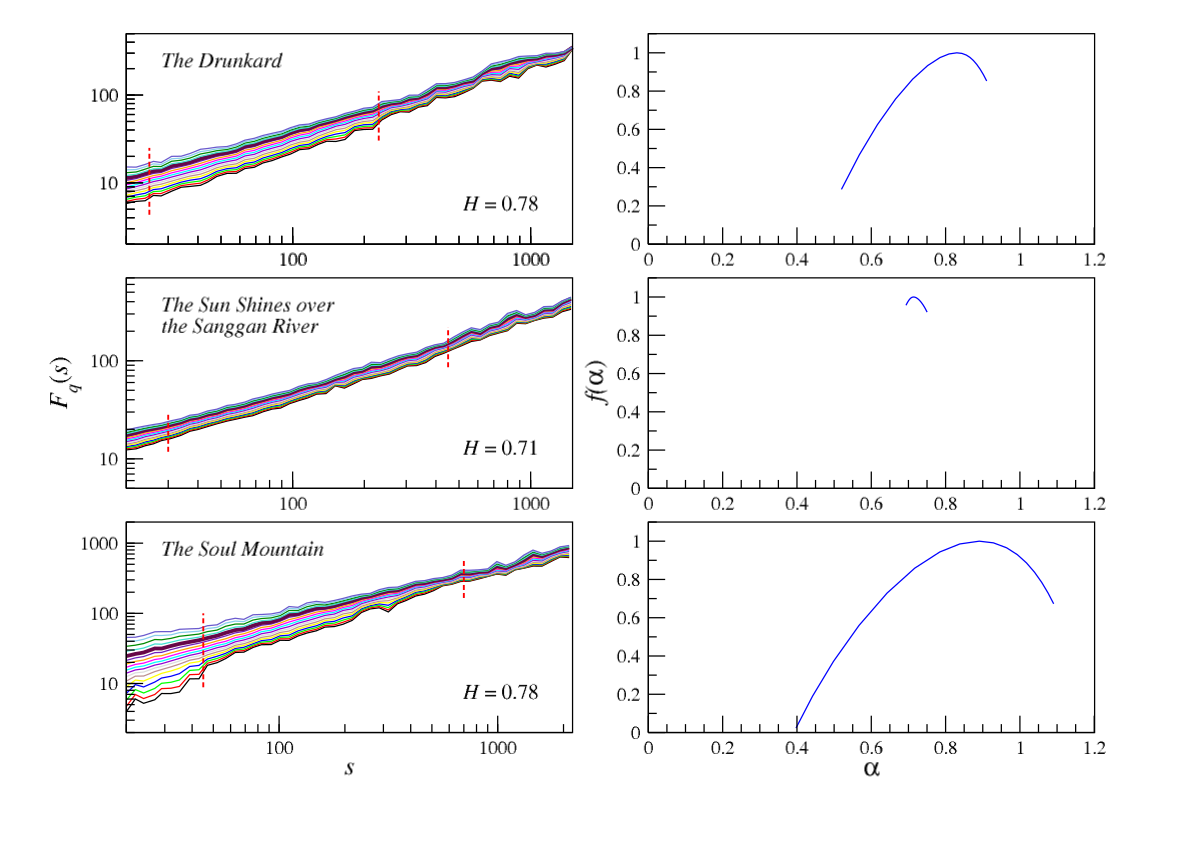}
\caption{The same quantities as in Fig.~\ref{fig::Fq.sentence.lengths.words} but for the sentence lengths measured in characters.}
\label{fig::Fq.sentence.lengths.characters}
\end{figure*}

This picture changes little if characters instead of words are chosen as the measurement units -- see Fig.~\ref{fig::Fq.sentence.lengths.characters}. The only difference is the lack of the scaling-range crossover in the novel \textit{The Sun Shines over the Sanggan River}. However, this novel shows clearly monofractal properties, while the other two novels show rich multifractality. The left-side asymmetry of $f(\alpha)$ for the latter stems from the fact that multifractality is associated in this case with large and medium-size fluctuations, while the small ones behave like monofractals~\cite{DrozdzS-2015a}. Values of the Hurst exponents are close to each other with $0.7 \le H \le 0.8$. Such values larger than 0.5 prove that the length of the consecutive sentences shows long-range autocorrelation.

\begin{figure*}
\centering
\includegraphics[width=1.0\textwidth]{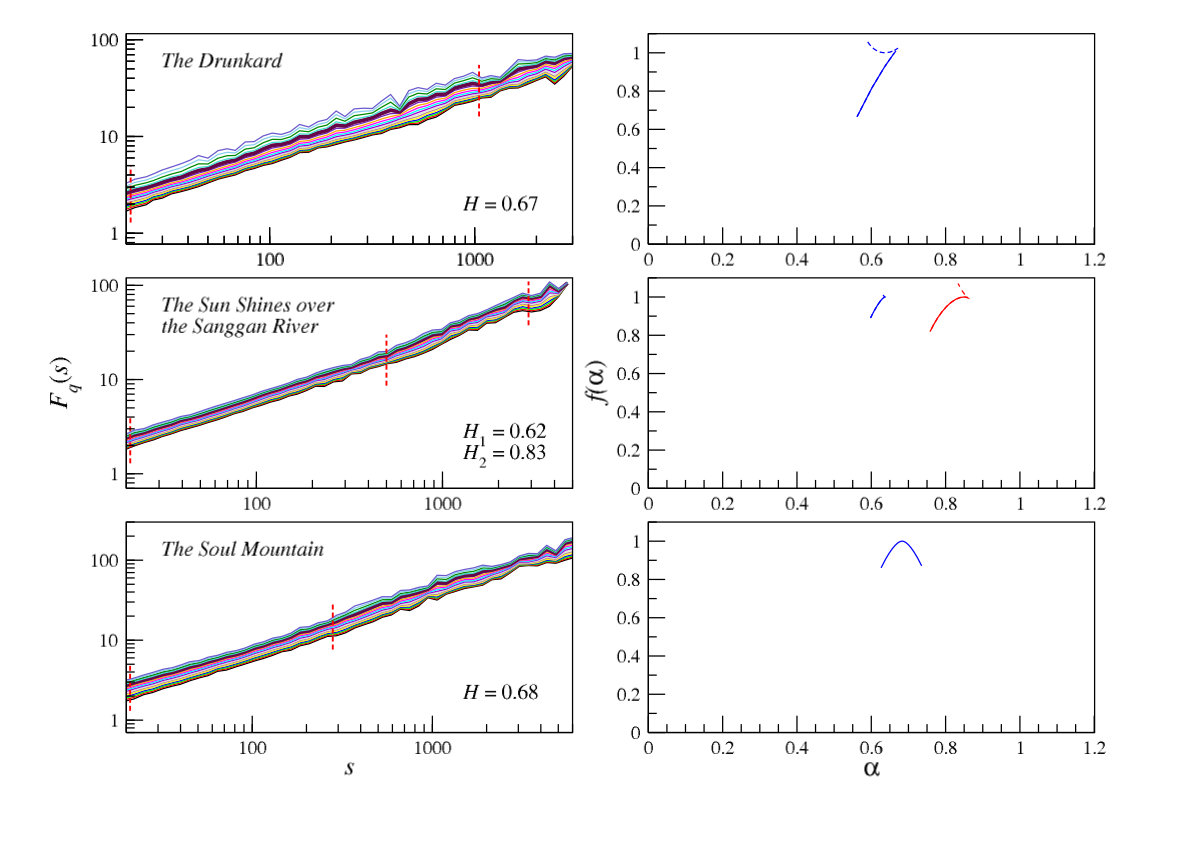}
\caption{A family of the fluctuation functions $F_q(s)$ with $-4 \le q \le 4$ (left) and the corresponding singularity spectra $f(\alpha)$ (right) calculated for time series of distances between consecutive punctuation marks measured in words for three Chinese novels: \textit{The Drunkard} (top), \textit{The Sun Shines over the Sanggan River} (middle), and \textit{The Soul Mountain} (bottom). In the last case, the largest observation, which can be viewed as an outlier, has been removed from the respective time series. Vertical dashed red lines in each plot on the left show the range of approximately power-law dependence of the fluctuation functions. In the case of \textit{The Sun Shines...} two such ranges have been identified, which results in two singularity spectra shown on the right. Non-physical behavior of $f(\alpha)$ is denoted by the dashed-line parts of the spectra.}
\label{fig::Fq.punctuation.mark.distance.words}
\end{figure*}

\begin{figure*}
\centering
\includegraphics[width=1.0\textwidth]{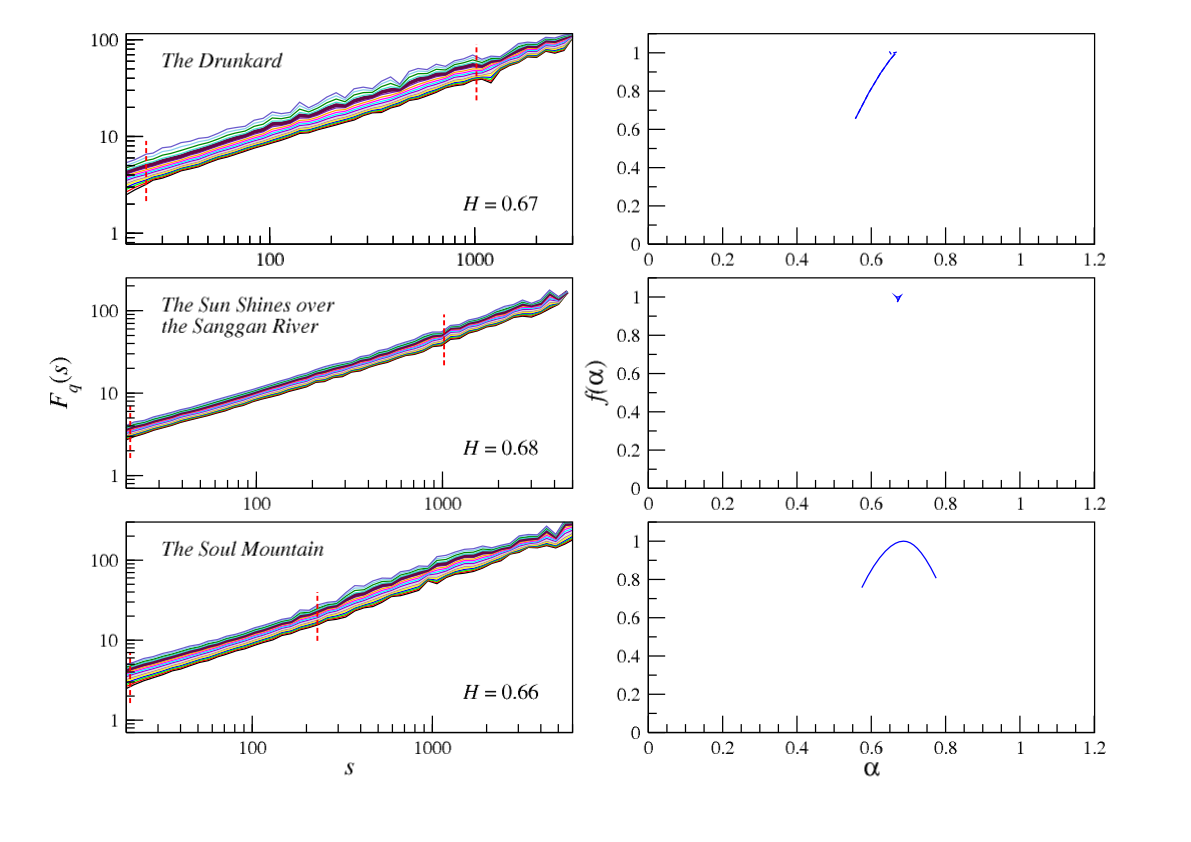}
\caption{The same quantities as in Fig.~\ref{fig::Fq.punctuation.mark.distance.words} but for the punctuation mark distances measured in characters.}
\label{fig::Fq.punctuation.mark.distance.characters}
\end{figure*}

The time series representing distances between consecutive punctuation marks have smaller amplitude than the ones representing sentence lengths and their CMFs have also thinner tails. This can have an impact on the fractal structure of the time series. In the case of the measurement performed in words, $F_q(s)$ for \textit{The Soul Mountain} and \textit{The Drunkard} show a single scaling regime, while there are two such regimes for \textit{The Sun Shines Over Sanggan River} -- this reproduces the situation seen in Fig.~\ref{fig::Fq.sentence.lengths.words}, but now the scaling worsens in the case of \textit{The Drunkard}. The Hurst exponents, while still indicating strong persistence, have slightly lower values for the two single-scaling-regime books ($H<0.7$) and exactly the same value for the third book with two scaling regimes. The singularity spectra are now much narrower and suggest only a monofractal scaling irrespective of their symmetry type. The existence of non-physical parts of the spectra with a convex shape indicates problems with scaling for certain values of $q$. An interesting situation appears for the distances between punctuation marks measured in characters -- Fig.~\ref{fig::Fq.punctuation.mark.distance.characters}. Here, each novel is represented by a single scaling range and the corresponding $f(\alpha)$ shows from none to only marginal amount of non-physical behavior. Surprisingly, \textit{The Soul Mountain} develops $f(\alpha)$ that may be considered multifractal ($\Delta\alpha > 0.2$) even though it was monofractal for the measurement in words in Fig.~\ref{fig::Fq.punctuation.mark.distance.words}. In contrast, the spectrum for \textit{The Sun Shines over the Sanggan River} is almost ideally point-like as for a theoretical monofractal. $f(\alpha)$ for \textit{The Drunkard} can also be considered as monofractal ($\Delta\alpha \approx 0.1$) despite the fact of its strong left-side asymmetry.

\begin{figure*}
\centering
\includegraphics[width=1.0\textwidth]{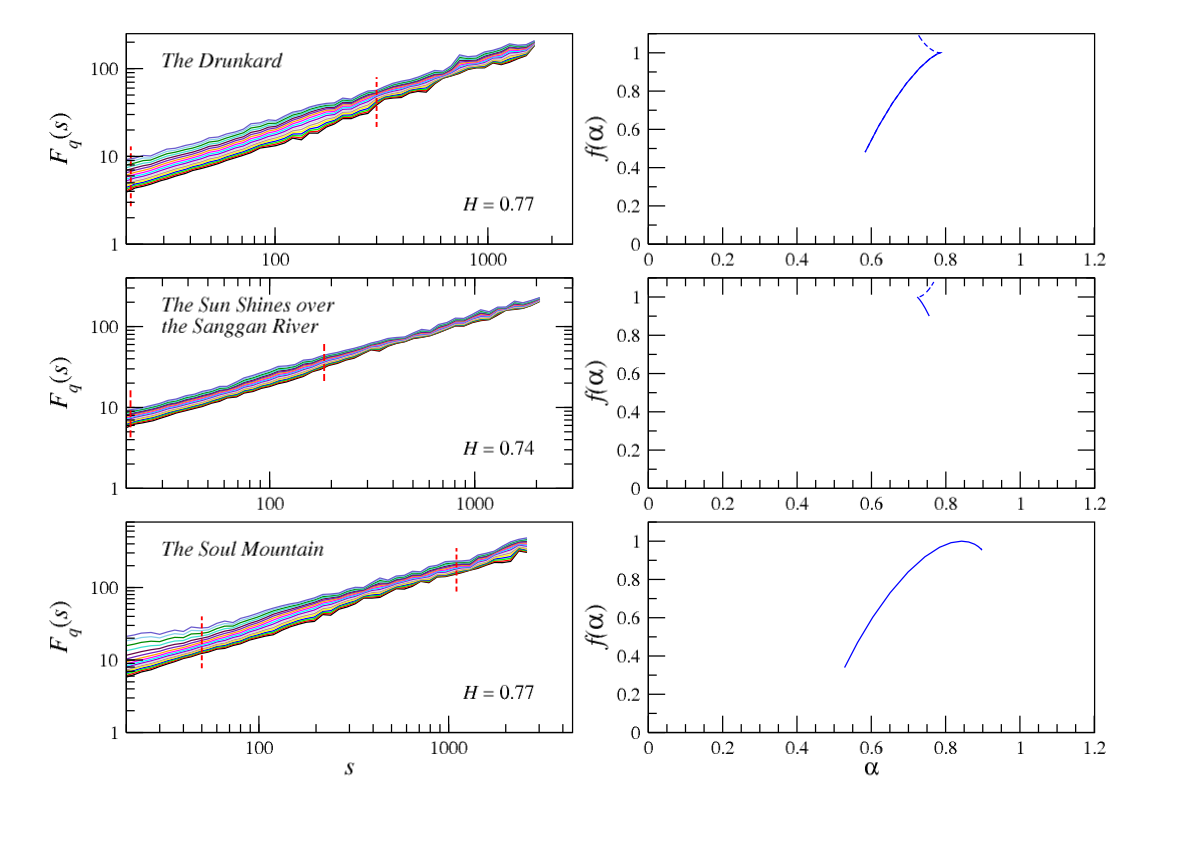}
\caption{A family of the fluctuation functions $F_q(s)$ with $-4 \le q \le 4$ (left) and the corresponding singularity spectra $f(\alpha)$ (right) calculated for time series of sentence lengths measured in words for English translations of three Chinese novels: \textit{The Drunkard} (top), \textit{The Sun Shines over the Sanggan River} (middle), and \textit{The Soul Mountain} (bottom).}
\label{fig::Fq.english.translations.sentence.lengths.words}
\end{figure*}

\begin{figure*}
\centering
\includegraphics[width=1.0\textwidth]{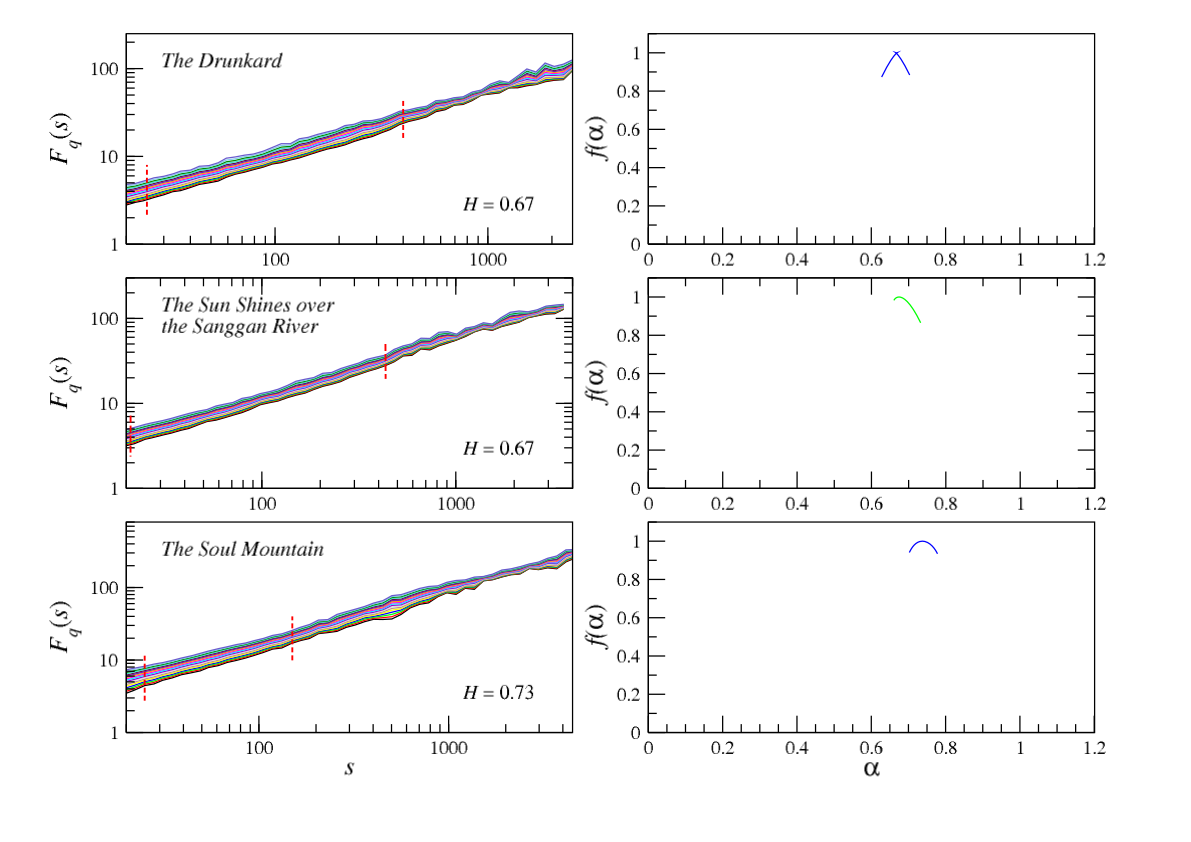}
\caption{The same quantities as in Fig.~\ref{fig::Fq.english.translations.sentence.lengths.words} but for the distances between consecutive punctuation marks.}
\label{fig::Fq.english.translations.punctuation.mark.distances.words}
\end{figure*}

Finally, one has to look at the fluctuation functions calculated for the time series of sentence lengths in the books translated into English. As before, because of the alphabetic writing system of English, the case of the measurement performed in characters will be omitted. If the units are words, the fluctuation functions calculated for the sentence lengths and displayed in Fig.~\ref{fig::Fq.english.translations.sentence.lengths.words} reveal only a single scaling range (with good power-law scaling) for each book with the Hurst exponent $H \approx 0.75$. These values are close to the values reported for the Chinese sources in Fig.~\ref{fig::Fq.sentence.lengths.words}. The associated singularity spectra are multifractal for \textit{The Drunkard} ($\Delta\alpha \approx 0.4$) and \textit{The Soul Mountain} ($\Delta\approx 0.2$) and monofractal for \textit{The Sun Shines over the Sanggan River} ($\Delta\alpha < 0.05$). This does not mean, however, that the translation process has not affected the fractal structure of the works, because it has introduced non-physical values of $f(\alpha)$ for \textit{The Drunkard} and \textit{The Sun Shines over the Sanggan River}. If the inter-mark distances are analyzed instead of the sentence lengths (Fig.~\ref{fig::Fq.english.translations.punctuation.mark.distances.words}), the picture changes significantly as all the books have monofractal spectra ($\Delta\alpha < 0.1$). This is in agreement with the results obtained for the Chinese texts, however (except for the fact that now no text show a scaling crossover in $F_q(s)$). The Hurst exponents $H \approx 0.7$ are slightly smaller but nevertheless are close to the respective values for the Chinese texts as well. It can be noted that the similarity between the source and translated texts in terms of multifractality was observed earlier for some other linguistic time series -- see Ref.~\cite{AusloosM-2012a}.

\section{Conclusions}

As a language that belongs to a different family than the Indo-European languages and whose grammar and writing system differs profoundly from the Western languages, Chinese constitutes an attractive object for a comparative study of its statistical properties, including the properties of its punctuation. In the present work, three aspects of literary Chinese language were studied: the word and punctuation-mark frequencies, the distribution of the distances between consecutive punctuation marks and the distribution of the sentence lengths, as well as the fractal properties of these quantities. The three novels that were considered show clear Zipf-like rank-frequency plots with the scaling index $\gamma \approx 1$ if the $n$-grams that constitute Chinese words are identified and counted. In full analogy to Western languages, the inclusion of punctuation marks into a Zipfian analysis improves scaling in the low-rank part of the Zipf plots that traditionally is modeled by the Zipf-Mandelbrot distribution. Values of the scaling index occur not to be invariant under translation to English and may change significantly. However, even in this case, the punctuation marks improve the agreement with a power law if included in the analysis.

The results show that both the distances between consecutive punctuation marks and the sentence lengths have distributions that can be modeled by the discrete Weibull distribution with the punctuation distances offering a slightly better agreement with the model. This does not differ from the results obtained earlier for sample books written in one of the Western languages~\cite{StaniszT-2023a,StaniszT-2024a}. If the Chinese texts are compared with their English translations, the fitted discrete Weibull distributions for the Chinese sources have larger $\beta$ (PMFs have thinner tails) than their translations. This can be a general property of these two languages, as the texts written in English originally (i.e., not those translated into English from Chinese) show similar behavior.

Recall at this point that the discrete Weibull distribution is equally effective in modeling the distributions of inter-punctuation distances in other major Western languages~\cite{StaniszT-2023a}, but, importantly, with parameters that largely distinguish these languages. The related robustness of this functional form can be taken as a strong indication that the discrete Weibull distribution is appropriate for describing such text characteristics regardless of the language used. These observations may help establish a parallel between biological processes related to survival where the effectiveness of Weibull distributions is known~\cite{MillerJrRG-1998a} and the cognitive, neurological and respiratory processes in the human body that necessitate the use of punctuation. In this regard more freedom is given by the use of sentence-ending punctuation marks, because, for example, a sequence of words can be taken as a sentence and thus terminated by a period, or the same sequence can be used as a dependent clause, and marked with a comma. The distribution of distances between sentence-ending punctuation marks can thus reflect the preferences of the author of a given text. Indeed, other functionally different distributions are considered in the literature~\cite{YuleGH-1939,WilliamsCB-1940,SichelHS-1974} in the context of sentence length, and, consistently, the results presented here also show that in this case the Weibull distributions are not obeyed as rigorously as in the case of complete punctuation.

Out of two quantities analyzed in this work, this is the sentence lengths that show multifractal structure. The novels with this property are \textit{The Soul Mountain} and, to a lesser extent, \textit{The Drunkard}. The former was written in an internal-monologue form and the latter shows the stream-of-consciousness narrative adopted by the modern Chinese literature from the Western authors such as James Joyce and Virginia Woolf~\cite{LiuJ-2023a}. In contrast, the sentence lengths in \textit{The Sun Shines over the Sanggan River} show convincingly monofractal behavior, consistent with its realist form and objective description.

It has to be pointed out that the present analysis was performed based on only three Chinese texts, which can limit the generality of the drawn conclusions. A much broader study is required in order to verify the observations presented here, especially those ones on the agreement between the empirical rank-frequency distributions and the Zipf law and on the obtained values of the parameter $\beta$ of the discrete Weibull distribution applied to punctuation distances and sentence lengths. Also, more texts have to be processed in order to establish a connection between the stream-of-consciousness narrative and multifractality of sentence length fluctuations. An earlier study of 95 Chinese texts from different epochs showed that about 75\% of them were monofractal, while only 12\% were convincingly multifractal~\cite{LiuJ-2023a}. However, as no relation between multifractality and the stream-of-consciousness narrative was established there due to a poor sample of the respective texts, this issue still remains to be addressed in more detail.

\section*{Data Availability Statement}

The data that support the findings of this study are available from the corresponding author upon reasonable request.

\nocite{*}
\bibliography{bibliography_file}
\end{CJK*}
\end{document}